\documentclass[journal]{IEEEtran}
\usepackage{times}
\usepackage{lineno}
\usepackage{amsmath}
\usepackage{amssymb}
\usepackage{commath}
\usepackage{tabularx}
\usepackage{booktabs}
\usepackage{float}
\usepackage{algorithm}
\usepackage{algpseudocode}
\usepackage{graphicx}
\usepackage{subfigure}
\usepackage{color}
\usepackage{CJK}
\usepackage{cite}
\usepackage{multirow}

\begin{document}
\title{Evaluating Local Geometric Feature Representations for 3D Rigid Data Matching}
\author{Jiaqi~Yang, Siwen~Quan, Peng~Wang and Yanning~Zhang,~\IEEEmembership{Senior Member, IEEE}
	
\thanks{This work is supported in part by the National Natural Science Foundation of China (No. 61876152). (\textit{Corresponding author: Jiaqi Yang}.)
	
	Jiaqi~Yang, Peng~Wang and Yanning~Zhang are with the National Engineering Laboratory for Integrated Aero-Space-Ground-Ocean Big Data Application Technology, School of Computer Science, Northwestern Polytechnical University, Xi'an 710129, China (e-mail: yjq\_hust@163.com; peng.wang@nwpu.edu.cn; ynzhang@nwpu.edu.cn).
	
	Siwen Quan is with the School of Electronic and Control Engineering, Chang'An University, Xi'an 710064,	China (e-mail: annequan@126.com).
	}
}
\markboth{Journal of \LaTeX\ Class Files,~Vol.~14, No.~8, August~2015}%
{Shell \MakeLowercase{\textit{et al.}}: Bare Demo of IEEEtran.cls for IEEE Journals}

\maketitle

\begin{abstract}
Local geometric descriptors remain an essential component for 3D rigid data matching and fusion. The devise of a rotational invariant local geometric descriptor usually consists of two steps: local reference frame (LRF) construction and feature representation. Existing evaluation efforts have mainly been paid on the LRF or the overall descriptor, yet the quantitative comparison of feature representations remains unexplored. This paper fills this gap by comprehensively evaluating nine state-of-the-art local geometric feature representations. Our evaluation is on the ground that ground-truth LRFs are leveraged such that the ranking of tested feature representations are more convincing as opposed to existing studies.  The experiments are deployed on six standard datasets with various application scenarios (shape retrieval, point cloud registration, and object recognition) and data modalities (LiDAR, Kinect, and Space Time) as well as perturbations including Gaussian noise, shot noise, data decimation, clutter, occlusion, and limited overlap. The evaluated terms cover the major concerns for a feature representation, e.g., distinctiveness, robustness, compactness, and efficiency. The outcomes present interesting findings that may shed new light on this community and provide complementary perspectives to existing evaluations on the topic of local geometric feature description. A summary of evaluated methods regarding their peculiarities is also presented to guide real-world applications and new descriptor crafting.
\end{abstract}


\IEEEpeerreviewmaketitle
\section{Introduction}\label{sec:intr}
\IEEEPARstart{T}{he} era of pervasive 3D technologies has arrived~\cite{rusu20113d}, with 3D data acquisition systems ranging from classical LiDAR scanner to commodity depth sensors such as Microsoft Kinect and Space Time cameras. Similar to the trend for 2D image fusion~\cite{james2014medical,ghassemian2016review}, great research efforts have characterized the realm of 3D rigid data (e.g., point clouds and meshes) matching and fusion (a.k.a, registration) with a number of real-world applications such as point cloud registration~\cite{rusu2008aligning,yang2016fast}, 3D reconstruction~\cite{yang2018aligning}, 3D object recognition~\cite{tombari2010unique,guo2013rotational}, localization~\cite{tateno20162}, and cultural heritage~\cite{gomes20143d}. Methods for 3D rigid data matching and fusion can be categorized as {\textit{local}} and {\textit{global}}. The arguably most well-known local method is the iterative closest points (ICP)~\cite{besl1992method} algorithm, but it suffers from the limitation of converging to the local minimum frequently without good initialization. For global methods, correspondences-based approaches~\cite{yang2017multi,guo2015novel,tombari2010object} have become a de-facto solution for many applications~\cite{guo20143d}. The establishment of point-to-point correspondences between 3D rigid shapes critically relies on the quality of the local geometric feature descriptor, whose goal is to comprehensively represent the contained geometric and spatial information within a local surface by a feature vector.

\begin{figure}[t]
	\centering
	\includegraphics[width=1\linewidth]{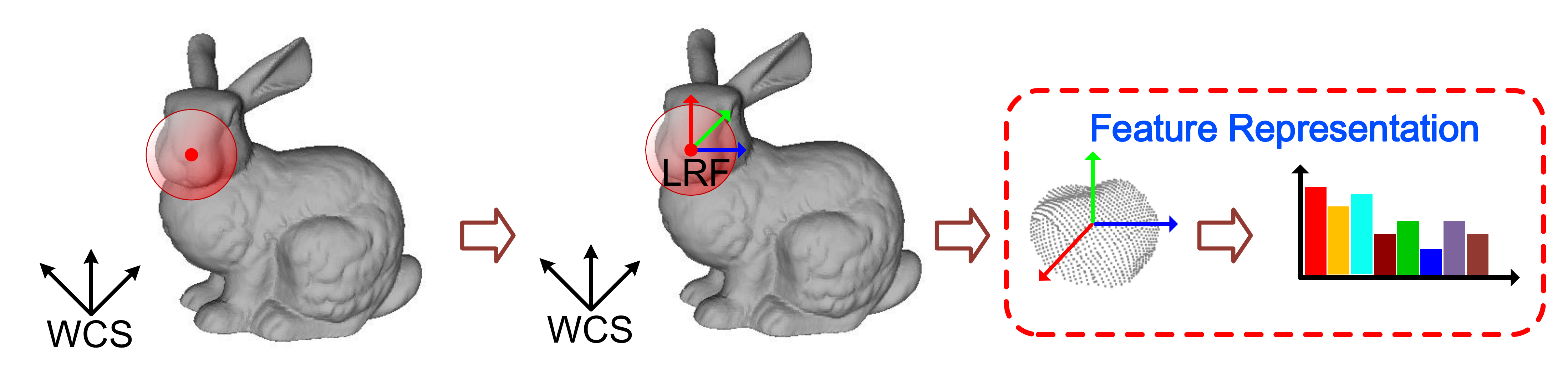}\\
	\caption{Illustration of local geometric feature representation. Left: a local surface (data within the red sphere) around a keypoint (the red dot) on a 3D rigid model. Middle: the construction of LRF to achieve independence from the world coordinate system (WCS). Right: encoding the geometric and spatial information in the transformed local surface with respect to the LRF  via a feature vector (the focus of this evaluation).}
	\label{fig:illus_eval}
\end{figure}

Usually,  a local reference frame (LRF) is first constructed in the local surface to on one hand make the descriptor rotational invariant and on the other encode the 3D spatial information such as to enhance the distinctiveness of the descriptor. Examples include signatures of histograms of orientations (SHOT)~\cite{tombari2010unique}, rotational contour signatures (RCS)~\cite{yang2017RCS_jrnl}, and local voxelized structure (LoVS)~\cite{Quan2018Local}. There are also some local geometric descriptors without LRF, e.g., spin images~\cite{johnson1999using} and 3D shape context (3DSC)~\cite{frome2004recognizing}. However, LRF-based descriptors have generally surpassed those LRF-independent ones on most publicly available datasets~\cite{guo2016comprehensive} and the LRF-based feature representation approaches have a vaster devising corpus. Despite traditional local feature descriptors, there are also some deep-learning-based descriptors such as PointNet~\cite{Charles2017PointNet}, 3DMatch~\cite{zeng20173dmatch}, CGF~\cite{khoury2017learning}, PPFNet~\cite{deng2018ppfnet}, PPF-FoldNet~\cite{Deng2018PPF}, and 3DFeat-Net~\cite{yew20183dfeat}. Unfortunately, most of existing learned descriptors are still sensitive to rotation~\cite{Deng2018PPF} and thus may limit their applications in real-world scenarios. Although CGF and PPF-FoldNet are rotational invariant, both descriptors take traditional feature representations to parametrize the input. Given above analysis,  this paper concentrates on traditional LRF-based feature representations (Fig.~\ref{fig:illus_eval}). 

To demonstrate the significance of our evaluation, we are amenable to answer the following questions.
\\
\\{\noindent{\textbf{1) Why are traditional feature representations still important?}}} Admittedly, deep-learning-based descriptors have outperformed many traditional local geometric descriptors in 2.5D scene registration context~\cite{zeng20173dmatch,Deng2018PPF}. The majority of current learned descriptors from raw point cloud data suffer from the sensitivity to rotation. An effective solution to this problem in both rigid (e.g., CGF~\cite{khoury2017learning} and PPF-FoldNet~\cite{deng2018ppfnet}) and non-rigid (e.g.,~\cite{Liu2019alt} and~\cite{Wang2018Learning} ) fields is first using traditional feature descriptors for parametrization  and then employing convolutional neural networks to further boost the performance. Consequently, the performance of these rotational-invariant learned descriptors critically relies on the quality of their used traditional feature representations. In addition, traditional LRF-based descriptors are still very competitive on tasks such as multi-point-cloud registration~\cite{yang2018aligning} and 3D object recognition~\cite{buch2018local}. 
\\
\\{\noindent{\textbf{2) Why have not existing evaluations addressed this concern?}}}  As illustrated in Fig.~\ref{fig:illus_eval}, the devise of a local descriptor usually contains two steps, i.e., LRF establishment and feature representation. Existing evaluations~\cite{guo2016comprehensive} mainly concentrate on the overall descriptors. This can tell us which descriptors are better but with limited insights to know which part makes it better.  Explaining this is critical for assessing the quality of point attributes, e.g., normals~\cite{tombari2010unique}, point densities~\cite{guo2013rotational,guo2015novel}, and signed distance~\cite{malassiotis2007snapshots,yang2017toldi}, for feature representation. For the evaluation of LRF, some studies have already been conducted~\cite{petrelli2011repeatability,petrelli2012repeatable,yang2018toward}. But no study yet, to the best of our knowledge, has been conducted to  specifically assess the quality of existing feature representations. The goal of this evaluation is to fill this gap and provide complementary insights to existing evaluations of LRFs~\cite{yang2018toward} and descriptors~\cite{guo2016comprehensive}.
\\
\\{\noindent{\textbf{3) What benefits can this evaluation bring to the community?}}} This paper will consider nine state-of-the-art feature representations and abstract their core  computational steps and ideas. It will therefore help the researchers to understand existing literature more deeply and devise either solid improvements based on existing methods or ground-breaking new approaches. In addition, the quantitative comparison makes it clear to the community about the trait of each feature representation. Based on current evaluations for LRFs~\cite{petrelli2011repeatability,petrelli2012repeatable,yang2018toward}, new descriptors can be proposed by simply replacing more proper feature representations under existing frames. Finally, this evaluation is beneficial for deep-learning-based descriptors because more reliable choices can be made regarding the selection of rotation invariant feature representations as the input for learning, rather than empirical selection~\cite{khoury2017learning}.

Motivated by above considerations, this paper first abstracts the core ideas and computational steps of nine state-of-the-art local geometric feature representations and then comprehensively evaluates their performance on six standard datasets. The  feature representations considered in this paper are mainly based on their popularity and state-of-the-art performance. \textbf{Because these methods are originally based on different LRFs, we specifically use ground-truth LRFs in this evaluation to eliminate the effect of LRF calculation errors and  independently assess the quality of different feature representations.} Note that we will still consider the effect of LRF errors in experiments because LRFs are not guaranteed to be fully repeatable especially in point cloud registration and object recognition applications~\cite{yang2018toward}, but such errors are made identical for all feature representations to ensure a fair comparison.   Our evaluation covers four major concerns for a feature representation. First, {\textit{distinctiveness}} is tested on experimental datasets with different data modalities (i.e., LiDAR, Kinect, and Space Time)  and application contexts (i.e., shape retrieval, point cloud registration, and object recognition). Second, {\textit{robustness}} is thoroughly tested with respect to a rich variety of nuisances including Gaussian noise, shot noise, data decimation, clutter, occlusion, and limited overlap. Third, {\textit{compactness}} is assessed by taking both storage and feature matching performance into consideration on all datasets. Forth, {\textit{computational efficiency}} regarding different scales of local surface is evaluated. Based on the experimental outcomes, we summarize the peculiarities, merits, and demerits of selected methods for evaluation. In a nutshell, this paper has three main contributions.
\begin{itemize}
	\item An abstraction of nine state-of-the-art local geometric feature representations regarding their core ideas and computational steps, which may help researchers to highlight the blind spots in this realm and devise new feature descriptors.
	\item A comprehensive evaluation of several feature representations in terms of  distinctiveness, robustness, compactness, and computational efficiency on six datasets with various data modalities, application scenarios, and nuisances. This evaluation presents complementary perspectives to~\cite{guo2016comprehensive,yang2018toward} on the topic of local geometric feature description.
	\item A summary of the peculiarities, advantages, and shortcomings of selected methods for evaluation that will provide instructive information to the developers in real-world applications and are inspiring to the following researchers.
\end{itemize}

The remainder of this paper is structured as follows. Sect.~\ref{sec:related} gives a review of 3D local geometric descriptors and relevant evaluations. Sect.~\ref{sec:mtd} presents the taxonomy and description of selected feature representations for evaluation. Sect.~\ref{sec:eval_mtd} introduces the experimental datasets, challenges, metrics, and implementation details. Sect.~\ref{sec:result} reports the experimental results with necessary discussions and explanations. Sect.~\ref{sec:sum} gives a summary of tested methods with respect to their traits, merits, and demerits. The conclusions are finally drawn in Sect.~\ref{sec:conc}.

\section{Related Work}\label{sec:related}
This section presents a brief review of existing local geometric feature representations and relevant evaluations on the topic of local feature descriptors for 3D rigid data.
\subsection{Local Geometric Feature Representations}
{\noindent{\textbf{Traditional methods.}}} Classical local geometric descriptors are either based on an LRF or not. As we concentrate on the feature representation stage, we will give more details to the feature representations of traditiaonal local geometric descriptors and ignore their employed LRFs (if used). A comprehensive overview of existing LRFs is available in~\cite{yang2018toward}.

For LRF-independent methods, spin images (SI)~\cite{johnson1999using} projects the neighboring points of a keypoint on a 2D plane by calculating their horizontal and vertical distances with respect to the tangent plane of the keypoint normal, and the ratio of points in each 2D grid is taken as the bin value. Local surface patches (LSP)~\cite{chen20073d}, similar to SI, projects 3D points to 2D as well but the projection is conducted in a latent feature space. Point feature histograms (PFH)~\cite{rusu2008aligning} and Fast PFH (FPFH)~\cite{rusu2009fast} leverage the point pair features extracted from the local surface to generate statistical histograms, where PFH considers all possible point pairs but FPFH speeds up the procedure by requiring each point pair to include the keypoint. Local feature statistics histograms (LFSH)~\cite{yang2016fast} fuses three histograms of normal deviation, signed distance, and point densities into a single feature vector. Due to the lack of spatial information, above features usually suffer from limited descriptiveness~\cite{guo2013rotational}.

For LRF-based methods, point signatures (PS)~\cite{chua1997point} first intersects the surface with a sphere to extract boundary points of the local surface and then computes the distance of the boundary points to the tangent plane of the keypoint normal in a clockwise order. Snapshots~\cite{malassiotis2007snapshots} first projects the local surface on a 2D plane in the LRF and then splits the 2D map into uniform grid; the feature value of each grid is the minimum signed projection distance of the points within the grid to the 2D plane. Later, triple orthogonal local depth images (TOLDI)~\cite{yang2017toldi} improves snapshots by computing three orthogonal local depth images to achieve a more comprehensive feature characterization. SHOT~\cite{tombari2010unique} divides the local 3D volume into a set of subspaces and concatenates the histograms of normal deviation in each subspace as the final feature representation. Unique shape context (USC)~\cite{tombari2010usc} partitions the local spherical volume along the radial, azimuth, and elevation directions and computes the point density for each partition as the feature value. Rotational projections statistics (RoPS)~\cite{guo2013rotational} proposes a ``rotation and projection'' mechanism that continually rotates the local surface with respect to the LRF and performs 3D-to-2D projection for each rotated surface; the eventual feature is the integration of the statistical information of the projected maps. Similar features based on ``rotation and projection'' include rotational contour signatures (RCS)~\cite{yang2017RCS_jrnl} and rotational silhouette maps (RSM)~\cite{Quan2018Representing}, where contour and silhouette cues of the 2D projection map are encoded, respectively.  Triple spin images (TriSI)~\cite{guo2015novel} is a variant of SI~\cite{johnson1999using} that concatenates three spin images calculated with respect to the three axes of the LRF, showing a great improvement over SI. Unlike the spherical volume used in most features, signatures of geometric centroids (SGC)~\cite{Tang2016Signature} and LoVS~\cite{Quan2018Local} instead employ a cubic volume to determine the local surface and they perform uniform spatial partition in the cubic volume to generate a set of voxels. SGC encode each voxel by the centroid and number of the points in the voxel, while LoVS simply uses a binary code to judge whether the voxel is empty or not. The leverage of spatial information for LRF-based methods greatly enhances their performance, yet, the quality of each feature representation remains unclear due to the lack of a comprehensive benchmarking comparison.
\\\\
{\noindent{\textbf{Deep-learning-based methods.}}} 3DMatch~\cite{zeng20173dmatch} is a pioneer learning-based representation that uses truncated distance function (TDF) to parametrize input local patches and learns the feature representation using a Siamese network paired with a metric learning network. CGF~\cite{khoury2017learning} first parametrizes the local surface with a traditional feature representation and then utilizes a multi-layer perception (MLP) network as  the feature embedding to shorten the initial feature as well as boost its distinctiveness. PPFNet~\cite{deng2018ppfnet} employs the point pair features between the keypoint and its neighbors to encode the raw local point cloud and proposes an N-branch network for feature learning. PPF-FoldNet~\cite{deng2018ppfnet} further improves PPFNet by leveraging rotational invariant point pair features and uses a point cloud auto-encoder network to achieve unsupervised feature learning. 3DFeat-Net~\cite{yew20183dfeat} takes the whole point cloud as  input and leverages an attention-based mechanism to simultaneously learn keypoints and descriptors. Deep-learning-based methods show great improvements over traditional methods in scene registration scenario, but most of them cannot generalize well and are sensitive to rotation~\cite{deng2018ppfnet}, limiting their effectiveness in real-world applications.
\subsection{Relevant Evaluations}
Sukno et al.~\cite{hansch2014comparison} performed a performance evaluation of local geometric descriptors for craniofacial landmarks on a set of point clouds for clinical study. Kim and Hilton~\cite{kim2013evaluation} presented a comparison of four local geometric descriptors for the registration of multi-modal data. Guo et al.~\cite{guo2016comprehensive} presented a quantitative evaluation of ten local geometric feature descriptors from different perspectives. This evaluation is closely related to this work but it pays attention to the overall feature descriptor rather than the particular feature representation.  Buch et al.~\cite{buch2016local} evaluated the feature matching and object recognition performance when utilizing multiple different local descriptors for fusion. Yang et al.~\cite{yang2017eval} quantitatively evaluated the effect of encoding different spatial information on the distinctiveness and robustness of local geometric features.

There are also some evaluation studies for the LRF. Petrelli and Stefano~\cite{petrelli2011repeatability} gave a study of LRF errors on the feature matching performance of local descriptors and evaluated the repeatability performance of seven LRFs in partial shape matching context. They later proposed a new metric to more accurately test the repeatability of an LRF and evaluated two additional LRFs on a variety of datasets~\cite{petrelli2012repeatable}. In addition to the concern of repeatability, Yang et al.~\cite{yang2018toward} presented an extensive evaluation of eight LRFs regarding their repeatability, robustness, efficiency, and feature matching performance when equipped with identical feature representations on six datasets. 

Above evaluation studies either focus on the LRF or the overall descriptor, yet the overall quality of existing feature representations remains unclear.

\section{Considered Feature Representations}\label{sec:mtd}
\begin{table}[t]\scriptsize
	\centering
	\caption{Notations.}
	\label{tab:notation}
	\scalebox{1}{
		\begin{tabular}{|c| c|}
			\hline
			$\bf{p}$ & Generic 3D keypoint\\
			\hline
			${\cal V}_s$& The local spherical volume around 	$\bf{p}$ \\
			\hline
			${\bf{q}}_r$ & A radius neighbor of $\bf{p}$\\
			\hline
			${\cal V}_c$& The local cubic volume around 	$\bf{p}$ \\
			\hline
			${\bf{q}}_c$ & A cubic neighbor of $\bf{p}$\\
			\hline
			$\bf{n}(\bf{p})$ & Normal vector of $\bf{p}$\\
			\hline
			${\cal L}(\bf{p})$ & LRF at $\bf{p}$\\
			\hline
	\end{tabular}}
\end{table}
This section introduces the main ideas and computational stages of nine local geometric feature representations, i.e., SHOT~\cite{tombari2010unique}, USC~\cite{tombari2010usc}, RoPS~\cite{guo2013rotational}, TriSI~\cite{guo2015novel}, SGC~\cite{Tang2016Signature}, TOLDI~\cite{yang2017toldi}, RCS~\cite{yang2017RCS_jrnl}, LoVS~\cite{Quan2018Local}, and RSM~\cite{Quan2018Representing}. We choose these representations based on their popularity and state-of-the-art performance. According to the taxonomy in~\cite{guo20143d}, they are either based on histogram or signature. We note that \textit{the abbreviations used for each methods here and hereinafter indicate the feature representation of the descriptor rather than the overall descriptor}, and in the following description for each method we assume that LRFs are already available (we will use ground-truth LRFs in the experiments).

\begin{figure}[t]
	\centering
	\includegraphics[width=1.0\linewidth]{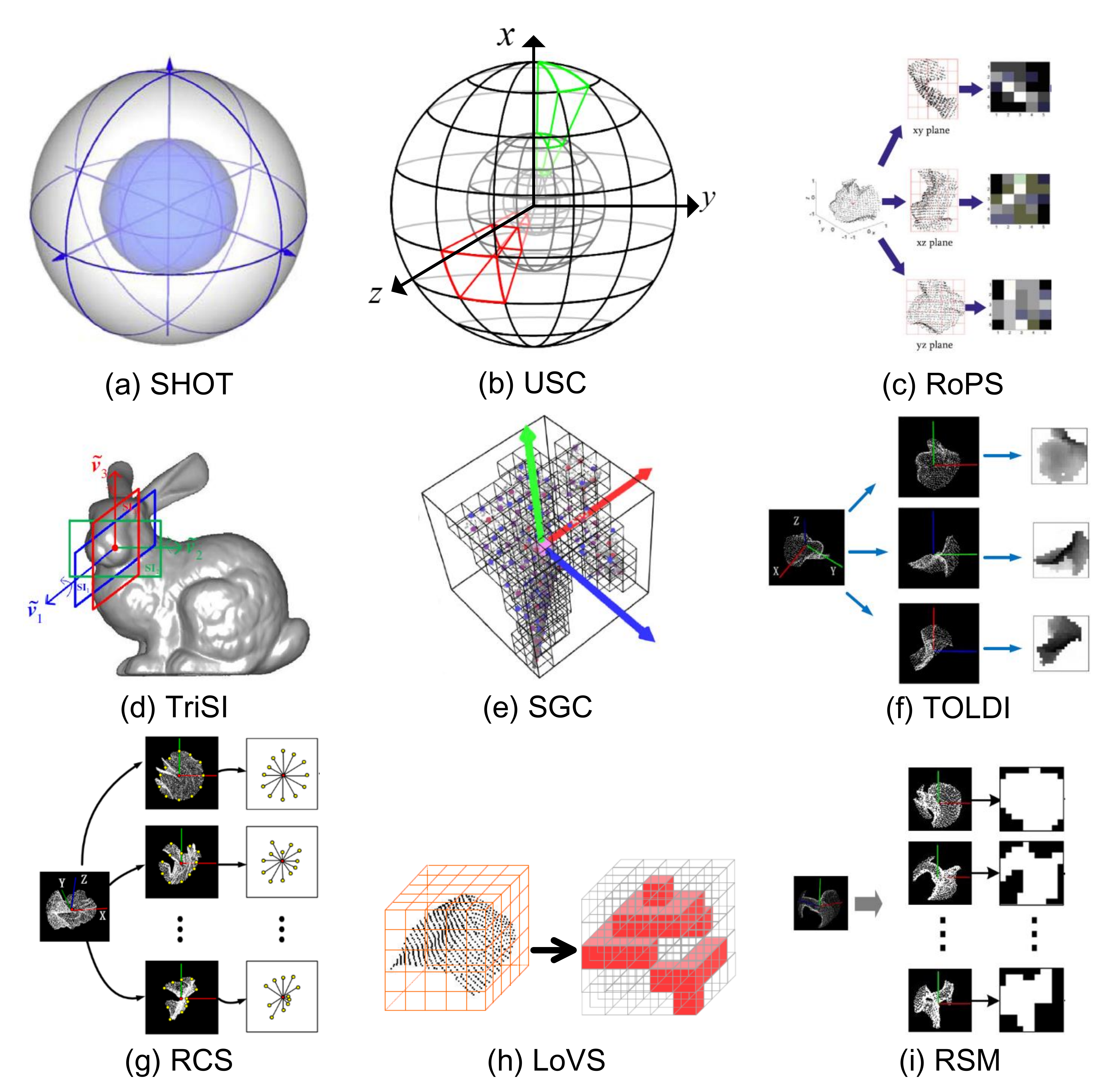}\\
	\caption{Schematic illustration of considered feature representations.}
	\label{fig:method}
\end{figure}

To help better understanding each feature representation, a schematic illustration of each method is shown in Fig.~\ref{fig:method} and some notations used for method introduction are reported in Table~\ref{tab:notation}.

\subsection{Histogram-based Methods}
Histogram-based methods describe the local surface through spatial distribution histogram, geometric attribute histogram, or a hybrid of them with spatial/geometric cues such as point numbers and surface normals.
\\\\
{\noindent{\textbf{SHOT~\cite{tombari2010unique}}}.} It takes hint from the SIFT~\cite{lowe2004distinctive} image descriptor that first splits the whole local volume into a set of subspaces and then calculates an attribute histogram for each subspace.

First, 	${\cal V}_s$ is divided into $N_{shot}^{div}$ sub-volumes along the radial, azimuth, and elevation axes. Then, for each point in a particular sub-volume, the cosine of the deviation angle between its normal and the $z$-axis of ${\cal L}(\bf{p})$ is calculated that will be later used for binning over the cosine space. Using cosine value instead of the deviation angle also speeds up the attribute extraction process. Finally, an attribute histogram with $N_{shot}^{bin}$ bins is computed for each sub-volume and all histograms are concatenated as the $N_{shot}^{div} \times N_{shot}^{bin}$-dimensional SHOT representation.
\\\\
{\noindent{\textbf{USC~\cite{tombari2010usc}}}.} Shape context is a well-known descriptor for 2D shape recognition~\cite{Belongie2002Shape}. USC extends it to 3D by collecting 3D point distribution information.

First, 	it splits ${\cal V}_s$ into bins with respectively $N_{usc}^L$ and $N_{usc}^K$ equally spaced boundaries in the azimuth and elevation directions but $N_{usc}^J$ logarithmically spaced boundaries in the radial direction. Neighboring points near to $\bf{p}$ are discarded to avoid being susceptive to small variations in shape close to the keypoint. Second, a weight $w({\bf{q}}_r)$ is calculated for each neighboring point ${\bf{q}}_r$ as:
\begin{equation}
w({{\bf{q}}_r}) = \frac{1}{{\rho \sqrt[3]{{V(j,k,l)}}}},
\end{equation}
where $j$, $k$, and $l$ represent the bin index of radial, elevation, and azimuth partitions, respectively; $V(j,k,l)$ is the volume of bin $(j,k,l)$; $\rho$ is the local point density of ${\bf{q}}_r$, i.e., the number of points in a small sphere centered at ${\bf{q}}_r$. The bin value is the accumulation of the weights of points falling inside the bin. By concatenating all bin values, the final $N_{usc}^K\times N_{usc}^L\times N_{usc}^J$-dimensional feature is obtained.
\\\\
{\noindent{\textbf{RoPS~\cite{guo2013rotational}}}.} This is the first method based on the ``rotation and projection'' mechanism, in oder to capture multi-view information represented by multiple 2D point distribution maps. RoPS is originally proposed for 3D object recognition~\cite{guo2013rotational} but also revealed to be very effective for surface registration~\cite{guo2014accurate}. 

Let ${\cal{N}}{(\bf p)}$ be the local surface inside ${\cal V}_s$. First, ${\cal{N}}{(\bf p)}$ is rotated around each coordinate axis of 	${\cal L}(\bf{p})$ by a set of discrete angles $\{\theta_1$,$\theta_2$,$\cdots$,$\theta_{N_{rops}^{rot}}\}$. Then, the rotate surface, e.g., ${\cal{N}}'{(\bf p)}$, are projected on the $xy$, $yz$, and $xz$ planes of the 	${\cal L}(\bf{p})$, creating three distribution maps that are further bounded by 2D rectangles. For each 2D distribution map, it is divided into ${N_{rops}^{div}}\times{N_{rops}^{div}}$ grids, and the central moment and Shannon entropy are employed to describe the distribution map. Specifically, the central moment $\mu_{mn}$ for matrix $\bf D$ with an order of $m+n$ is defined as:
\begin{equation}
{\mu _{mn}} = \sum\limits_{i = 1}^{N_{rops}^{div}} {\sum\limits_{j = 1}^{N_{rops}^{div}} {{{(i - \overline i )}^m}{{(j - \overline j )}^n}{\bf{D}}(i,j)} } ,
\end{equation}
where $\overline i  = \sum\limits_{i = 1}^{N_{rops}^{div}} {\sum\limits_{j = 1}^{N_{rops}^{div}} {i{\bf{D}}(i,j)} } $ and $\overline j  = \sum\limits_{i = 1}^{N_{rops}^{div}} {\sum\limits_{j = 1}^{N_{rops}^{div}} {j{\bf{D}}(i,j)} } $. The Shannon entropy $e$ is defined as:
\begin{equation}
e =  - \sum\limits_{i = 1}^{N_{rops}^{div}} {\sum\limits_{j = 1}^{N_{rops}^{div}} {{\bf{D}}(i,j)log({\bf{D}}(i,j))} }.
\end{equation}
RoPS finally chooses 5 combinations of these statistics and integrates the statistical information of all distribution maps for feature description with a dimensionality of $N_{rops}^{rot}\times 3 \times 3 \times 5$.
\\\\
{\noindent{\textbf{TriSI~\cite{guo2015novel}}}.} TriSI is a variant of the spin image descriptor~\cite{johnson1999using} that generates three spin  images around different spinning axes to achieve a comprehensive information description, showing great improvements over spin image~\cite{guo2016comprehensive}.

It maps 3D points to a 2D space represented by two parameters, i.e., $\alpha$ and $\beta$. In particular, the definitions of the two parameters are given as:
\begin{equation}
\left\{ \begin{array}{l}
\alpha  = \sqrt {||{{\bf{q}}_r} - {\bf{p}}|{|^2} - {{({\bf{v}} \cdot ({{\bf{q}}_r} - {\bf{p}}))}^2}} \\
\beta  = {\bf{v}} \cdot ({{\bf{q}}_r} - {\bf{p}})
\end{array} \right.,
\end{equation}
where $\bf v$ is chosen as the $x$-axis, $y$-axis, and $z$-axis of the LRF 	${\cal L}(\bf{p})$, respectively. By partitioning each projection with $N_{trisi}^{div}\times N_{trisi}^{div}$ grids, three spin images are generated. The concatenation of all spin images is the $3\times N_{trisi}^{div}\times N_{trisi}^{div}$-dimensional TriSI representation.
\subsection{Signature-based Methods}
The methods in this category compute one or more geometric attributes individually for each 3D point within the local support or each 2D grid after performing 3D-to-2D projection.
\\\\
{\noindent{\textbf{SGC~\cite{Tang2016Signature}}}.} Instead of using a local spherical volume, SGC proposes constructing a bounding cubical volume to simplify the extraction of local geometric features as well as the descriptor construction.

First, the local cubic volume is regularly partitioned into $N_{sgc}^{div}\times N_{sgc}^{div} \times N_{sgc}^{div}$ voxels. Second, the centroid $(X_c, Y_c,Z_c)$ and the number of the points $N_{pnts}$ within each voxel are utilized to represent the shape feature of a voxel. The motivation of extracting centroid feature is twofold. One is that centroid is an integral feature, making it robust to noise and point density variation~\cite{Pottmanna2009Integral}. The other is that the computation of centroid only requires linear computational operations. To save storage, the centroid feature is compressed into a single value via $C=(Z_c \times L+Y_c)\times L+X_c$, where $L$ is the  length of cube edges. The dimension of SGC  therefore is $2\times N_{sgc}^{div}\times N_{sgc}^{div} \times N_{sgc}^{div}$.
\\\\
{\noindent{\textbf{TOLDI~\cite{yang2017toldi}}.} TOLDI takes use of the signed distance cue to generate local depth images for the description of 3D local shape. It simultaneously achieves promising performance in shape retrieval, 3D object recognition, and point cloud registration scenarios~\cite{yang2017toldi}.
	
First, the local surface is projected with respect to the $xy$, $yz$, and $xz$ planes of the LRF 	${\cal L}(\bf{p})$, generating three orthogonal 2D maps. Each map is then divided uniformly into $N_{toldi}^{div}\times N_{toldi}^{div}$ grids. Unlike the popular point distribution map representation used in RoPS and TriSI, TOLDI chooses the minimum of the projection distances (a.k.a., local depth or signed distance) of those points within a 2D grid to the projection plane as the bin value. The reason of opting the minimum projection distance is imitating the humans' vision mechanism and occluded information cannot be captured. The integration of three orthogonal local depth images forms the final feature with $3\times N_{toldi}^{div}\times N_{toldi}^{div}$ dimensions.
\\\\
{\noindent{\textbf{RCS~\cite{yang2017RCS_jrnl}}.} It presents the first attempt to employ 2D contour information for the description of 3D geometric information. To remedy the limitation that contour, though shown to be robust to noise, has limited discriminative information, it borrows the idea from RoPS and employs multi-view contour information.

First, the local surface is  rotated $N_{rcs}^{rot}$ times around the three axes of ${\cal L}(\bf{p})$ simultaneously  by an incremental  angle. For each rotated surface, it is then projected on a 2D plane (the $xy$-plane of ${\cal L}(\bf{p})$). Subsequently, the intersection point of a ray and the projection map with the largest distance from $\bf p$ is defined as a contour point. Here, the first ray is aligned with the $x$-axis and the angular gap between any two neighboring rays is identical. A total of $N_{rcs}^{c}$ contour points are extracted after each rotation and the signature formed by concatenating the distances from $\bf p$ to the contour points is calculated for representing a rotated surface. The concatenation of all contour signatures is the eventual $N_{rcs}^{rot}\times N_{rcs}^{c}$ feature representation.
\\\\
{\noindent{\textbf{LoVS~\cite{Quan2018Local}}.} This method, similar to SGC, also encodes the local geometric information within a cubical volume. The difference is that LoVS discards the description for the geometric information inside a voxel, it instead judges the occupancy of each voxel to craft a binary feature.

The calculation of LoVS is straightforward. It first splits the local cubic volume into $N_{lovs}^{div}\times N_{lovs}^{div} \times N_{lovs}^{div}$ uniform voxels. A voxel is labeled as 1 if there are points inside; otherwise, 0. These labels are concatenated in a pre-defined order, generating the LoVS representation with $N_{lovs}^{div}\times N_{lovs}^{div} \times N_{lovs}^{div}$ bits.
\\\\
{\noindent{\textbf{RSM~\cite{Quan2018Representing}}.} This representation is the first to leverage silhouette images for the characterization of local shapes. The binary nature of silhouette makes RSM binary as well. It demonstrates that simply using silhouette images can achieve comparable distinctiveness as point distribution and local depth images, yet being even more robust.

The calculation of RSM is very similar to RoPS and RCS, i.e., rotation and projection are frequently performed to gain multi-view information. For each 2D $N_{rsm}^{div}\times N_{rsm}^{div}$ projection map, RSM converts it to a silhouette image through judging if there are points insider a grid. In order to improve the robustness to outliers, a \textit{connectivity} constraint is enforced on each grid. A total of $N_{rsm}^{rot}$ silhouette images are generated and concatenated, generating a feature with $N_{rsm}^{rot} \times N_{rsm}^{div}\times N_{rsm}^{div}$ bits.
\section{Evaluation Methodology}\label{sec:eval_mtd}
\begin{figure}[t]
\centering
\includegraphics[width=1.0\linewidth]{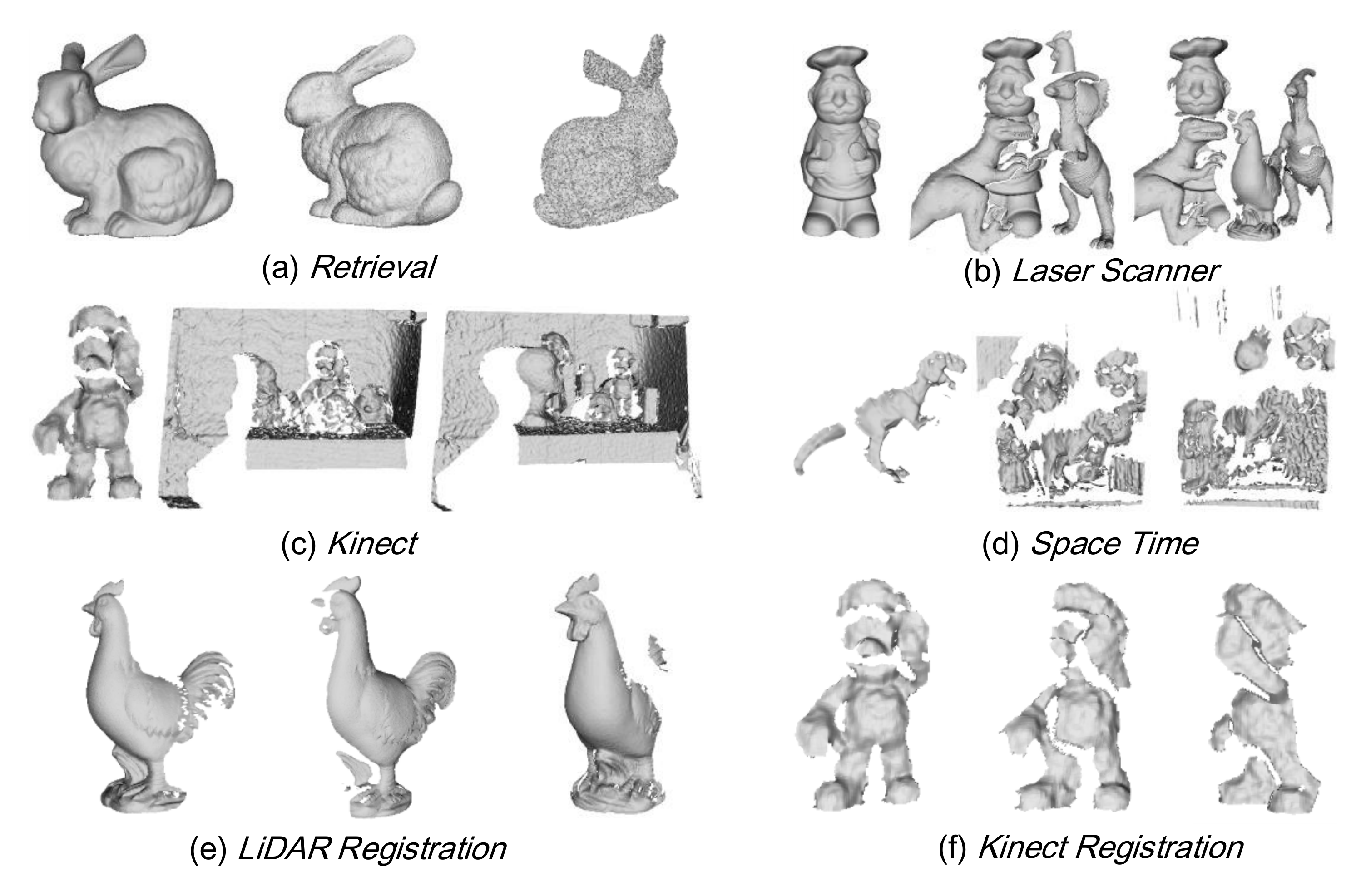}\\
\caption{Six datasets for experimental evaluation. For each dataset, samples of a source model and two target models are visualized.}
\label{fig:dataset}
\end{figure}
In this section, datasets and  metrics for experimental evaluation are presented. The nuisances offered by each dataset or manually injected  are also detailed to test a feature representation's robustness. Finally, implementation details of evaluated methods are introduced. In the following, we will use \textit{source} and \textit{target} models to denote two rigid point clouds to be matched.
\subsection{Datasets}
{\noindent{\textbf{Retrieval~\cite{tombari2013performance}}.} It has three groups of matching pairs. The first group is a noise-free subset while the other two groups contain two levels of Gaussian noise with standard deviations being 0.3 \textit{pr} and 0.5 \textit{pr}, respectively. The unit \textit{pr} here and hereinafter denotes the point cloud resolution, i.e., the average  of the distances from each point to its nearest neighbor. Each group has 18 matching pairs and the target models are the rotated copies of the source models. This dataset addresses the 3D shape retrieval scenario and the main challenge is Gaussian noise.
\\\\
{\noindent{\textbf{Laser Scanner~\cite{mian2006three,mian2010repeatability}}.} This dataset addresses model-based 3D object recognition scenario, which contains 5 source models and 50 target models with different levels of clutter and occlusion. A total of 188 valid matching pairs are available for this dataset. The point clouds were scanned using a laser scanner and incorporate uniformly distributed and dense points.
\\\\
{\noindent{\textbf{Kinect~\cite{tombari2010unique}}.} It is a 3D object recognition dataset and was collected with a Microsoft Kinect v1.0 sensor, consisting of 26 source models and 15 target models. It provides 43 matching pairs with challenges such as real noise, clutter, and occlusion for evaluation.
\\\\
{\noindent{\textbf{Space Time~\cite{tombari2013performance}}.} The \textit{Space Time} dataset addresses rigid matching in 3D object recognition scenario that was captured using space-time stereo technology. 6 source models and 12 models are provided by this dataset, forming 24 matching cases. Note that although it is a 3D object recognition dataset, the source models are 2.5D point clouds.
\\\\
{\noindent{\textbf{LiDAR Registration~\cite{mian2006three}}.} There are 22, 16, 16, and 21 2.5D views of four models in this dataset scanned by a laser scanner. It addresses 3D rigid registration scenario and offers 496 valid matching pairs having at least 10\% overlap (c.f. Eq.~\ref{eq:overlap} for the definition of overlap). The point clouds in this dataset have well-preserved structures. Challenges of this dataset include limited overlap and self-occlusion.
\\\\
{\noindent{\textbf{Kinect Registration~\cite{tombari2010unique}}.} It consists of 15, 16, 20, 13, 16, and 15 partial point cloud views of six object models. This dataset addresses 3D rigid registration scenario for low-quality point clouds, i.e., data acquired by a Microsoft Kinect v1.0 sensor. Analogous to the \textit{LiDAR Registration} dataset, we only consider matching pairs with at least 10\% overlap. This dataset is more challenging than the \textit{LiDAR Registration} dataset because it is additionally contaminated by real noise.
\\\\
The experimental datasets have the following peculiarities. (i) Varying application contexts. The matching of rigid data finds a wide range of real-word applications and the selected datasets address three typical application contexts, e.g., shape retrieval (the \textit{Retrieval} dataset), model-based 3D object recognition (the \textit{Laser Scanner}, \textit{Kinect}, and \textit{Space Time} datasets), and point cloud registration (the \textit{LiDAR Registration} and \textit{Kinect Registration} datasets). Thus, they offer challenges from different application contexts such as clutter, occlusion, and limited overlap for a thorough evaluation. (ii) Different data modalities. Data captured by LiDAR, Kinect, and space time acquisition systems are considered in our evaluation. Different sensor technology results in point clouds with various qualities. For instance, data generated by LiDAR consist of fairly dense points with good uniformity, while data acquired by Kinect and space-time systems often suffer from sparsity, real noise, and holes. Since the sensor may vary with applications and performance demands, it is desired to considering data with different modalities.
\subsection{Nuisances}
A desired  geometric feature representation should be robust to common nuisances. Some nuisances only exist for a particular application, e.g., clutter and occlusion in 3D object recognition; but some are independent from applications, e.g., noise. Therefore, we classify these common nuisances into two categories: application-dependent and application-independent. We follow a rule, i.e., independently examining the effect of one particular nuisance to avoid unclear impacts resulted by a mixture of nuisances. The nuisance-free matching case is that all matched local patches are identically fused. So we choose the first subset of B3R dataset (noise-free) where the target models are rotated copies of the source models and enforce all application-independent nuisances to the target models for robustness evaluation regarding application-independent perturbations. As for application-dependent nuisances, we choose the \textit{Laser Scanner} and \textit{LiDAR Registration} datasets to include challenges existed in 3D object recognition and point cloud registration scenarios. Although there are additional datasets to them, e.g., \textit{Kinect} and \textit{Kinect Registration}, these datasets include a mixed nuisances. e.g., noise and outliers, and will make it ambitious for evaluation. We also note that the raw \textit{Retrieval}  dataset  only concerns about noise so we do not consider shape retrieval datasets regarding the evaluation of robustness to  application-dependent nuisances. Details of each nuisance are given as follows.
\subsubsection{Application-dependent Nuisances}
~
\\
{\noindent{\textbf{Varying support radii}.} The support radius of a geometric feature representation, denoted by $R$,  determines its spatial scale. A small support radius will result in limited geometric information to be encoded, yet a large support radius may drag in unwanted perturbations such as data from cluttered objects. Thus, stay distinctive across difference scales of local surface patches is challenging for a feature representation. The scale issue is frequently investigated for 3D object recognition~\cite{guo2013rotational,tombari2013performance} so we test this term on the  \textit{Laser Scanner} dataset. We vary the support radius of each tested geometric feature from 5 \textit{pr} to 30 \textit{pr} with an interval of 5 \textit{pr}.
\begin{figure}[t]
\centering
\includegraphics[width=0.4\linewidth]{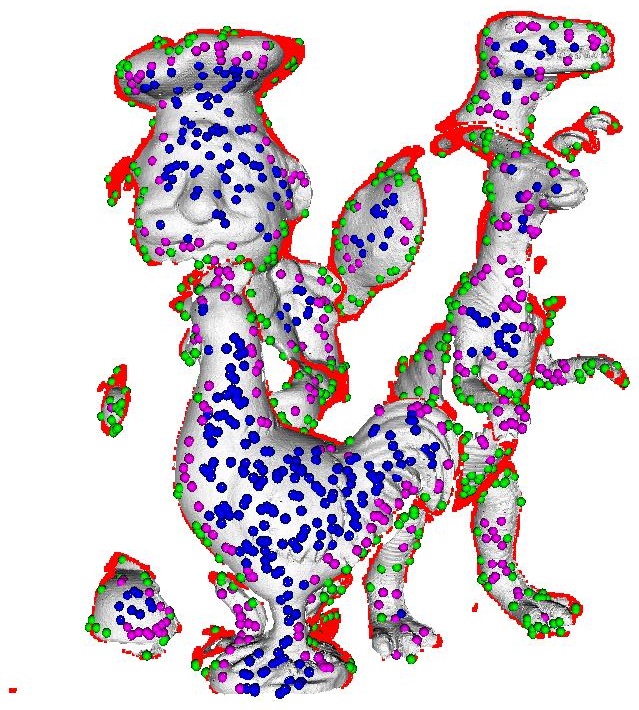}\\
\caption{Illustration of three sets of randomly sampled keypoints (the total number is 1000) on a scene from the \textit{Laser Scanner} dataset with different distances to the boundary. Red points: detected boundary using the method in~\cite{mian2010repeatability}; green , purple, and blue points represent keypoints whose distances to the boundary are in the range of $[0,0.5R)$, $[0.5R,1.0R)$ and $[1.0R,\infty)$, respectively.}
\label{fig:dist2bound_illus}
\end{figure}
\\\\
{\noindent{\textbf{Distance to boundary}.} Boundary effect caused by occlusion is troublesome because it will lead to data missing and corrupt the original local geometric structure. This is mainly owing to the limitation of current 3D data acquisition systems that they can only capture the 2.5D view data for a 3D object. To preserve sufficient numbers of correct feature correspondences, it is desired for a good geometric feature to retrieve correspondences in boundary regions. 

Therefore, we first  detect boundary for the target models in the \textit{Laser Scanner} dataset. Here we use the approach in~\cite{mian2010repeatability} for boundary detection in point clouds. Then, keypoints on the target models are then split into six groups in which each point distances to the boundary are in the ranges of $[0,0.2R)$, $[0.2R,0.4R)$, $[0.4R,0.6R)$, $[0.6R,0.8R)$, $[0.8R,1.0R)$, and $[1.0R,\infty)$, respectively. Fig.~\ref{fig:dist2bound_illus} gives an illustration of several sets of keypoints with different distances to the boundary (for clarity purpose, we show three sets of keypoints in this figure).
\\\\
{\noindent{\textbf{Clutter}.} Clutter refers to as additional data with respect to the object to be recognized in the target model. Without preprocessing such as segmentation, more candidates should be considered in the matching phase for each keypoint in the source model and many of them may have similar geometric structures to the true corresponding local patch, resulting in repeatable patterns. Specifically, clutter is defined as~\cite{mian2006three,guo2016comprehensive}:
\begin{equation}
{\rm clutter}=1-\frac{{\rm source}\;{\rm surface}\;{\rm area}\;{\rm in}\;{\rm target}}{{\rm total}\;{\rm surface}\;{\rm area}\;{\rm of}\;{\rm target}}.
\end{equation}

We accordingly divide the \textit{Laser Scanner} dataset  into 7 groups with less than 65\%, $[65\%,70\%)$, $[70\%,75\%)$, $[75\%,80\%)$, $[80\%,85\%)$, $[85\%,90\%)$, and $[90\%,95\%)$ clutter, respectively.
\\\\
{\noindent{\textbf{Occlusion}.} Cluttered objects in the front of the object to be recognized will cause occlusion. This is a long-standing issue for 3D rigid data matching~\cite{guo20143d}. Occlusion, according to~\cite{mian2006three,guo2016comprehensive},  is defined as:
\begin{equation}
{\rm occlusion}=1-\frac{{\rm source}\;{\rm surface}\;{\rm area}\;{\rm in}\;{\rm target}}{{\rm total}\;{\rm source}\;{\rm surface}\;{\rm area}}.
\end{equation}

We partition the \textit{Laser Scanner} dataset, similar to the test of robustness to clutter,  into 7 groups with less than 60\%, $[60\%,65\%)$ $[65\%,70\%)$, $[70\%,75\%)$, $[75\%,80\%)$, $[80\%,85\%)$, and $[85\%,90\%)$ of occlusion, respectively.
\\\\
{\noindent{\textbf{Partial overlap}.} Data pairs scanned from different viewpoints usually have only limited overlapping region. According to~\cite{mian2006novel}, the overlap ratio of two point clouds is given as:
\begin{equation}\label{eq:overlap}
{\rm overlap}=\frac{\#\;{\rm corr.}\;{\rm points}\;{\rm between}\;{\rm model} \;{\rm and} \;{\rm scene}}{{\rm min}(\#\;{\rm model}\;{\rm points},\#\;{\rm scene}\;{\rm points})}.
\end{equation}

This experiment is deployed on the \textit{LiDAR Registration} dataset. We hence split this dataset into 7 subsets with less than 0.3, $[0.3,0.4)$,  $[0.4,0.5)$,  $[0.5,0.6)$, $[0.6,0.7)$, $[0.7,0.8)$, and $[0.8,0.9)$ overlaps, respectively.
\subsubsection{Application-independent Nuisances}
\begin{figure}[t]
\centering
\includegraphics[width=1.0\linewidth]{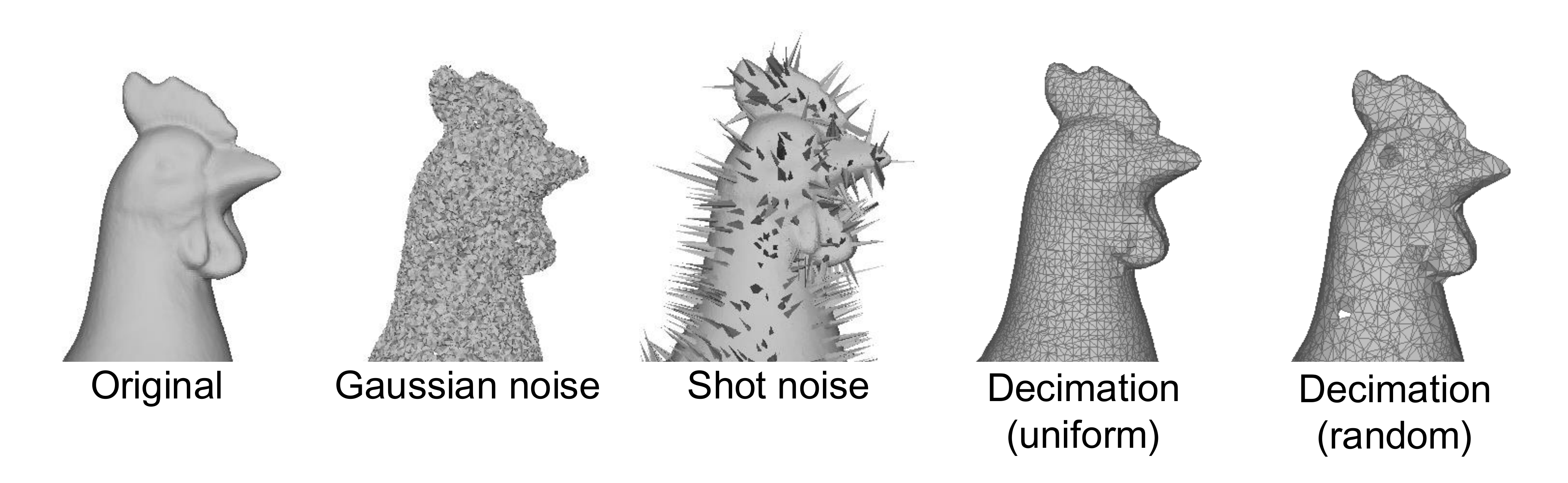}\\
\caption{Visual illustration of four application-independent nuisances including Gaussian noise (0.5 \textit{pr} standard deviation), shot noise (3\% outliers), uniform data decimation ($\frac{1}{8}$ decimation rate), and random data decimation ($\frac{1}{8}$ decimation rate).}
\label{fig:nuisance_view}
\end{figure}
~
\\
{\noindent{\textbf{Gaussian noise}.} Noise may appear as perturbations of points, or unwanted points close to a	3D surface~\cite{tam2013registration}. Eight levels of Gaussian noise whose standard deviations range from 0.25 \textit{pr} to 2 \textit{pr} with an interval of 0.25 \textit{pr} are respectively added to the target models. The noise is independently injected from the $x$-,$y$-, and $z$-axis of each point. A visual illustration of a model with  0.5 \textit{pr} Gaussian noise is shown in Fig.~\ref{fig:nuisance_view} (the second column).
\\\\
{\noindent{\textbf{Shot noise}.}  Shot  noise, i.e., outliers, are undesired points far from the 3D	surface. To synthetic shot noise, we follow the procedure in~\cite{zaharescu2012keypoints}. Specifically, we move a portion of points (i.e., $r_{shot}$ of the total number of points in the target model) along their normal vectors with a given distance ($d_{shot}$), because shot noise usually lie along the viewing direction and such way can well approximate real shot noise. Eight levels of shot noise are considered, i.e., $r_{shot}$ is set from 1\% to 8\% with an incremental step of 1\%. In this evaluation, $d_{shot}$ is set to 0.8$R$~\cite{yang2018toward}. The visual illustration of a model with 3\% shot noise is presented in Fig.~\ref{fig:nuisance_view} (the third column).
\\\\
{\noindent{\textbf{Data decimation}.} The  change of distance from the sensor to the object/scene will result in data resolution variation for 3D point clouds, as compared to the scale variation for images. Handling sparse data is challenging but necessary especially for remote sensing applications. We down-sample the target models to $\frac{1}{2}$, $\frac{1}{4}$, $\frac{1}{8}$, $\frac{1}{16}$, and $\frac{1}{32}$ of their original resolutions, respectively. In most prior works, the data used in such experiment are uniformly down-sampled~\cite{guo2013rotational,yang2017toldi}. Nevertheless, the change of data resolution may not in a uniform manner. We thus propose a new evaluation setup for this term, i.e., random down-sampling. As illustrated by Fig.~\ref{fig:nuisance_view} (the last two columns), the model after random down-sampling has lower quality than the one after uniform-down-sampling.  Besides, performing random down-sampling will generate data with irregularly distributed points. The term of point irregularity remains has been overlooked in past studies but is necessary for robustness evaluation.
\begin{figure}[t]
\centering
\includegraphics[width=0.6\linewidth]{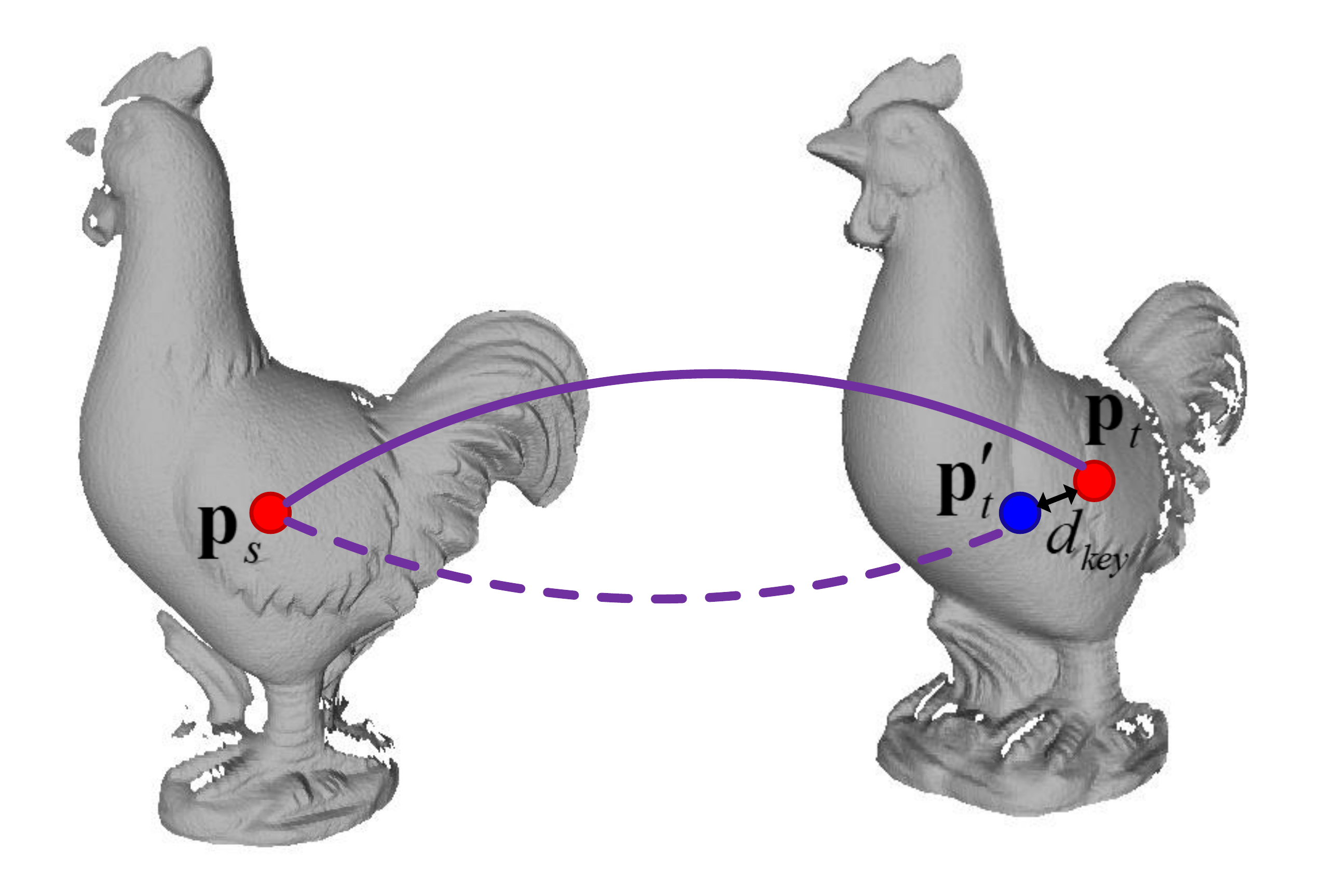}\\
\caption{Illustration of key point localization error. Points ${\bf p}_s$ and ${\bf p}_t$ are correctly matched without localization errors, whereas points ${\bf p}_s$ and ${\bf p}'_t$ suffer from a key point localization error of $d_{key}$.}
\label{fig:key_error}
\end{figure}
\\\\
{\noindent{\textbf{Keypoint localization error}.} Prior to  feature extraction and matching, a set of distinctive keypoints need to be detected. An evaluation of existing 3D keypoint detectors~\cite{tombari2013performance} reveals that the repeatability performance of current 3D keypoint detectors is still limited, so one has to expect keypoint localization errors. By default, the performance of a feature representation is evaluated on ground-truth keypoints to eliminate the effect of keypoint localization errors~\cite{guo2013rotational,tombari2010unique} (Sect.~\ref{subsec:metric}), thus enabling a specific evaluation for other terms. In this experiment, we focus on the impact of keypoint detection errors on feature matching.

Let $({\bf p}_s,{\bf p}_t)$ be a pair of corresponding keypoints between the source and target data. Then, a point ${\bf p}'_t$ is localized in the target model with a distance of $d_{key}$ to ${\bf p}_t$. Hence, a localization error $d_{key}$ is enforced on point pair $({\bf p}_s,{\bf p}'_t)$ (Fig.~\ref{fig:key_error}). This procedure is repeated for all ground-truth corresponding keypoint pairs and the performance is evaluated on these re-organized keypoints. In particular, we vary $d_{key}$ from 1 \textit{pr} to 6 \textit{pr} with a gap of 1 \textit{pr}, thus obtaining datasets with six levels of keypoint localization error. 
\\\\
{\noindent{\textbf{LRF error}.} As previously mentioned in Sect.~\ref{sec:intr}, we use ground-truth LRFs by default to specifically examine the quality of feature representations. However, existing  LRF methods are not guaranteed to be repeatable~\cite{petrelli2011repeatability,yang2018toward} and may produce different degrees of LRF errors in various cases. To address this concern, we examine the performance of geometric features when their LRFs suffer from different degrees of angular errors.

The LRF error is defined as the overall angular error between corresponding axes of two LRFs~\cite{petrelli2011repeatability,yang2018toward}. As suggested by~\cite{petrelli2011repeatability}, we specifically examine the effect of LRF errors injected from the $x$-axis, the $z$-axis, and both axes. The $y$-axis is not necessary to be considered because it is orthogonal to other axes. For each case, we vary the LRF error from 2.5 degrees to 15 degrees with an increasing step of 2.5 degrees. Note that when injecting LRF errors from two axes, the angular error is randomly split for both axes. 

\subsection{Metrics}\label{subsec:metric}
The distinctiveness of a geometric feature is usually quantitatively evaluated using the Recall versus 1-Precision Curve (RPC)~\cite{guo2013rotational,tombari2010unique,yang2017RCS_jrnl}. We also use this criterion for evaluation, which is defined as follows. First, a source feature is matched against all target features to determine the closest and second closest target features. If the ratio of the smallest distance to the second smallest distance is smaller than a threshold $\tau_{fm}$, the source feature and the closest target feature are identified as a match. Only if their associated ketpoints are spatially close enough, the matched is judged as correct. Assume that there are $N_{corr}$ correspondences, $N_{match}$ of them are defined as matches, and there are $N_{match}^{correct}$ correct matches in total, 1-Precision is defined as:
\begin{equation}
{\rm 1- Precision}=1- \frac{N_{match}^{correct}}{N_{match}}.
\end{equation}
Recall is defined as:
\begin{equation}
{\rm Recall}=\frac{N_{match}^{correct}}{N_{corr}}.
\end{equation}
By varying $\tau_{fm}$ from 0 to 1, a curve would be generated. As in~\cite{guo2013rotational,yang2017RCS_jrnl}, 1000 keypoints are randomly selected from the source model and their corresponding point are located in the target model via ground-truth information. RPC is then calculated by matching these keypoints.

For the comparison of features' robustness, we use the area under curve (AUC) metric to aggregately measure an RPC's quality~\cite{guo2016comprehensive}. By varying the test condition, e.g., adding different levels of nuisance, a curve will be generated.
The compactness of a feature representation, as in~\cite{yang2017eval}, is reflected by taking its AUC and storage into consideration simultaneously. We finally examine the computational efficiency by comparing the time cost required for each feature when extracting features for local surfaces with different scales.
\subsection{Implementation Details}
\begin{table}[t]\tiny
	\centering
	\caption{Parameter settings of evaluated methods.}
	\label{tab:para}
	\scalebox{1}{
		\begin{tabular}{|c|c| c|c|c|}
			\hline
			\bf	No.& \bf Features&\bf Parameters&\bf Dimensionality&\bf Data type\\
			\hline
			1&SHOT~\cite{tombari2010unique}&$N_{shot}^{div}=32, N_{shot}^{bin}=11$&352&Float\\
			\hline
			2&USC~\cite{tombari2010usc}&$N_{usc}^{K}=12, N_{usc}^{L}=11, N_{usc}^{J}=15$&1980&Float\\
			\hline
			3&RoPS~\cite{guo2013rotational}&$N_{rops}^{rot}=3$&135&Float\\
			\hline
			4&TriSI~\cite{guo2015novel}&$N_{trisi}^{div}=15$&675&Float\\
			\hline
			5&SGC~\cite{Tang2016Signature}&$N_{sgc}^{div}=8$&1024&Float\\
			\hline
			6&TOLDI~\cite{yang2017toldi}&$N_{toldi}^{div}=20$&1200&Float\\
			\hline
			7&RCS~\cite{yang2017RCS_jrnl}&$N_{rcs}^{rot}=6, N_{rcs}^{c}=12$&72&Float\\
			\hline
			8&LoVS~\cite{Quan2018Local}&$N_{lovs}^{div}=9$&729&Binary\\
			\hline
			9&RSM~\cite{Quan2018Representing}&$N_{rsm}^{rot}=6, N_{rsm}^{div}=11$&726&Binary\\
			\hline
	\end{tabular}}
\end{table}

All methods are implemented in the point cloud library (PCL) 1.8.1~\cite{rusu20113d}. SHOT, USC, RoPS are available in PCL;  the code of TOLDI, RCS, LoVS, and RSM was provided by the authors; TriSI and SGC are our re-implementations in PCL. The parameters of all methods are reported in Table~\ref{tab:para} (we use default parameter settings suggested by the authors). Because SHOT needs point normals for feature representation, we employ the principle component analysis (PCA)-based method~\cite{hoppe1992surface} for normal extraction where the neighborhood size is set to 20. The support radius for all descriptor, by default, is set to 15 \textit{pr}~\cite{guo2016comprehensive,yang2017RCS_jrnl}.

The experiments were conducted on a laptop with a 3.4 GHz CPU and 24 GB RAM without usages of parallel computing techniques or GPU implementation.
\section{Experimental Results}\label{sec:result}
In this section, experimental results produced by following the setup in Sect.~\ref{sec:eval_mtd}  along with relevant explanations and discussions are presented. 
\begin{figure}[htbp]
	\begin{minipage}{0.49\linewidth}
		\raggedleft
		\subfigure[\textit{Retrieval}]{
			\includegraphics[width=0.9\linewidth]{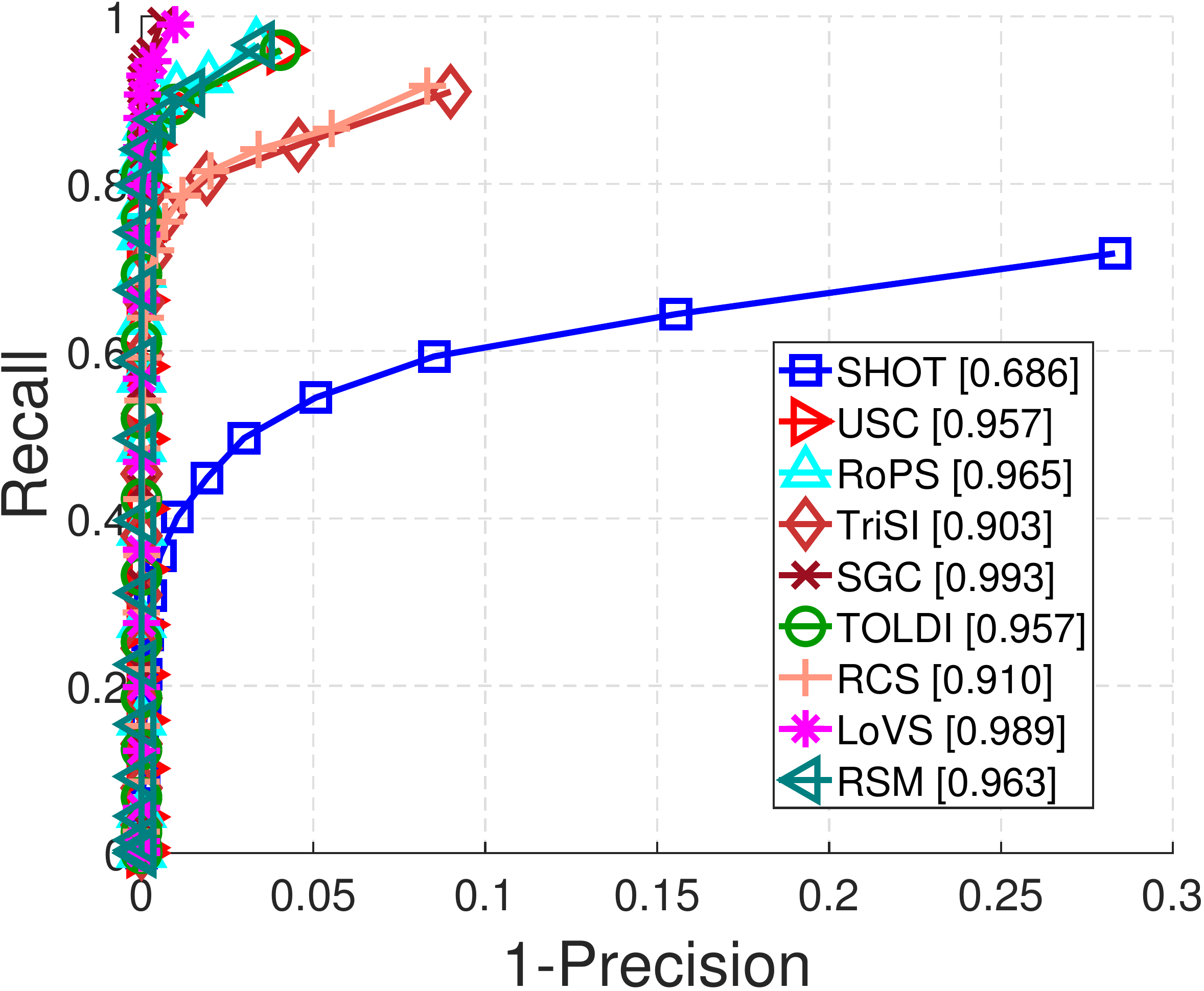}}
	\end{minipage}
	\begin{minipage}{0.49\linewidth}
		\raggedright
		\subfigure[\textit{Laser Scanner}]{
			\includegraphics[width=0.9\linewidth]{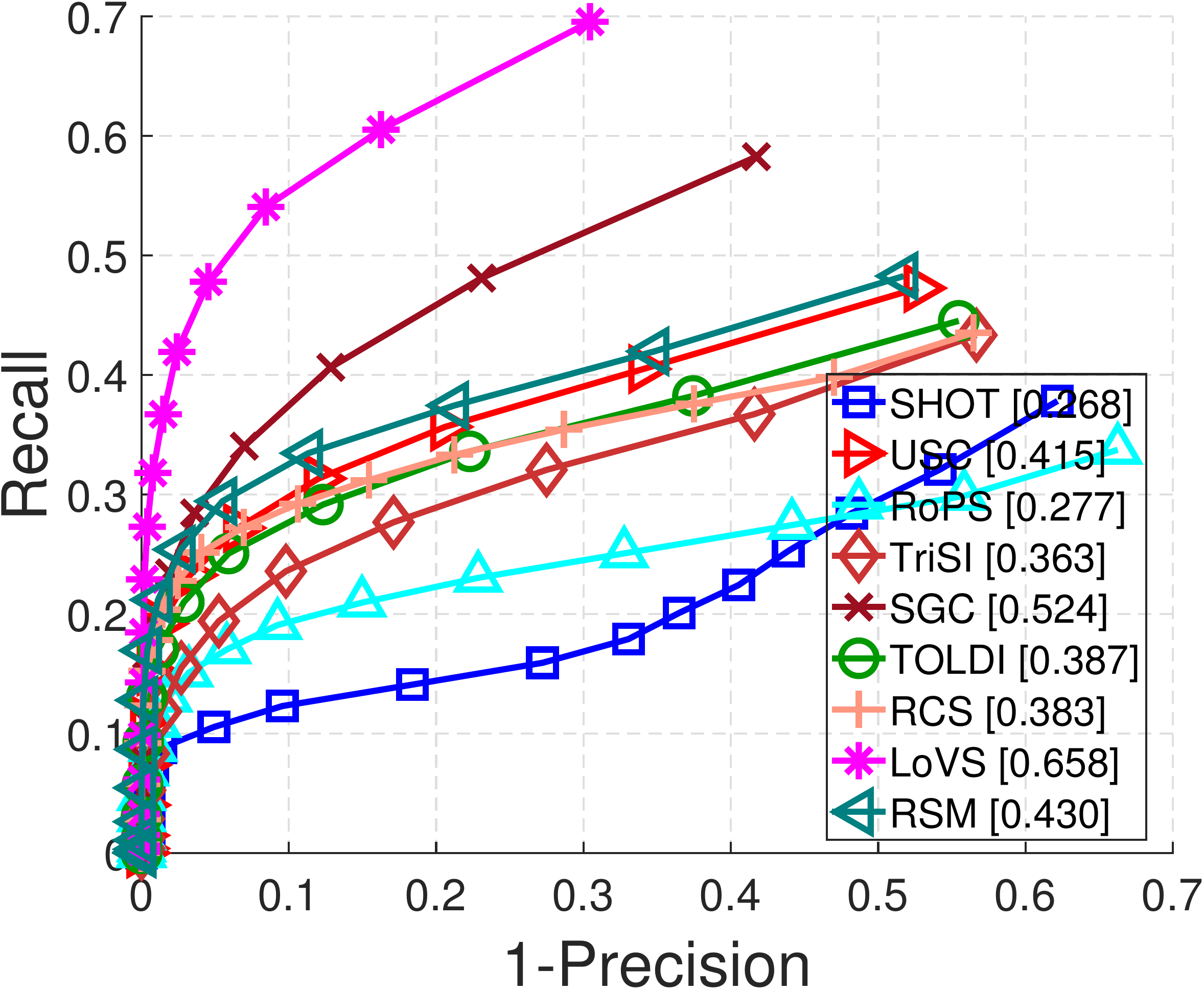}}
	\end{minipage}	
	\begin{minipage}{0.49\linewidth}
		\raggedleft
		\subfigure[\textit{Kinect}]{
			\includegraphics[width=0.9\linewidth]{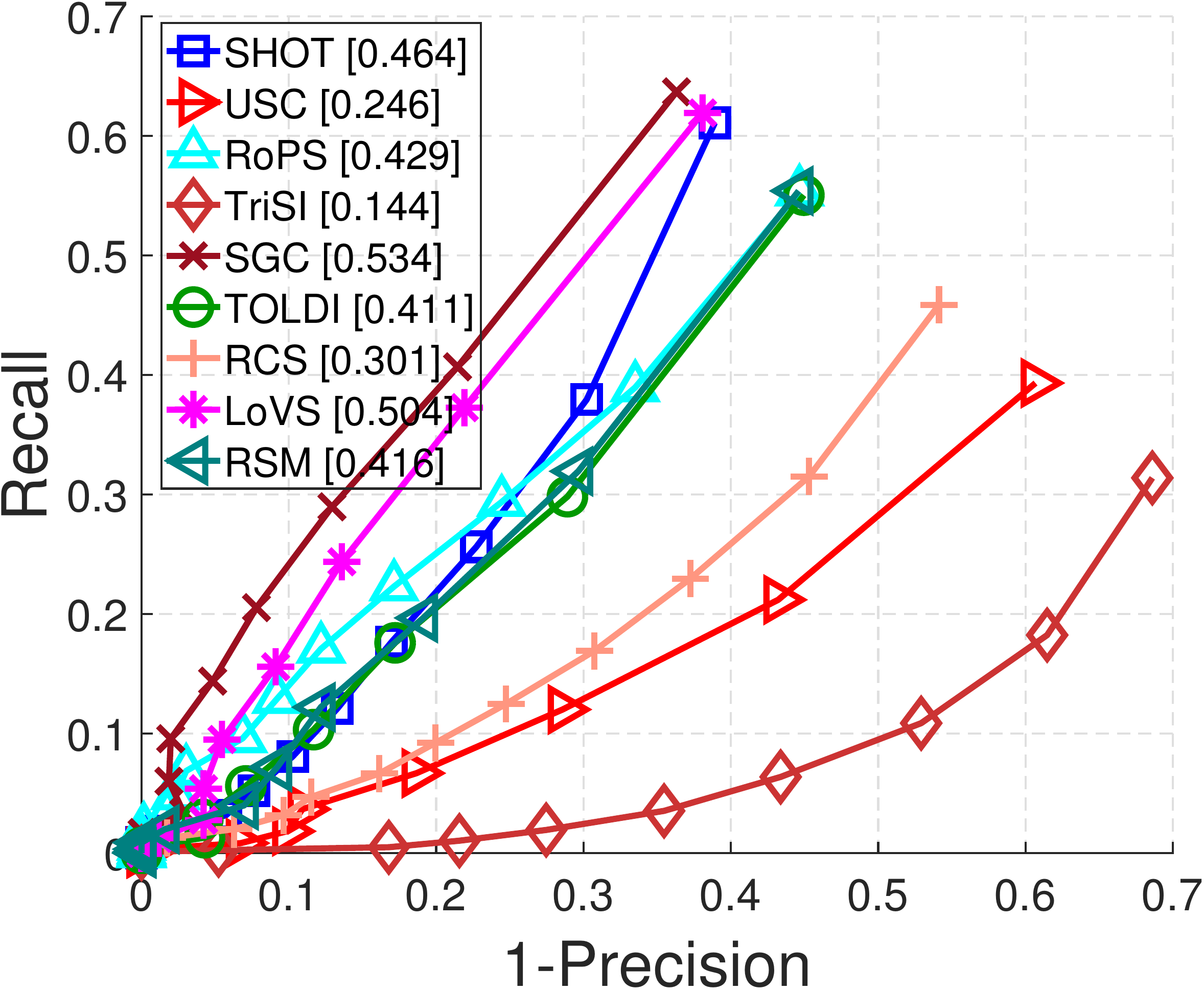}}
	\end{minipage}
	\begin{minipage}{0.49\linewidth}
		\raggedright
		\subfigure[\textit{Space Time}]{
			\includegraphics[width=0.9\linewidth]{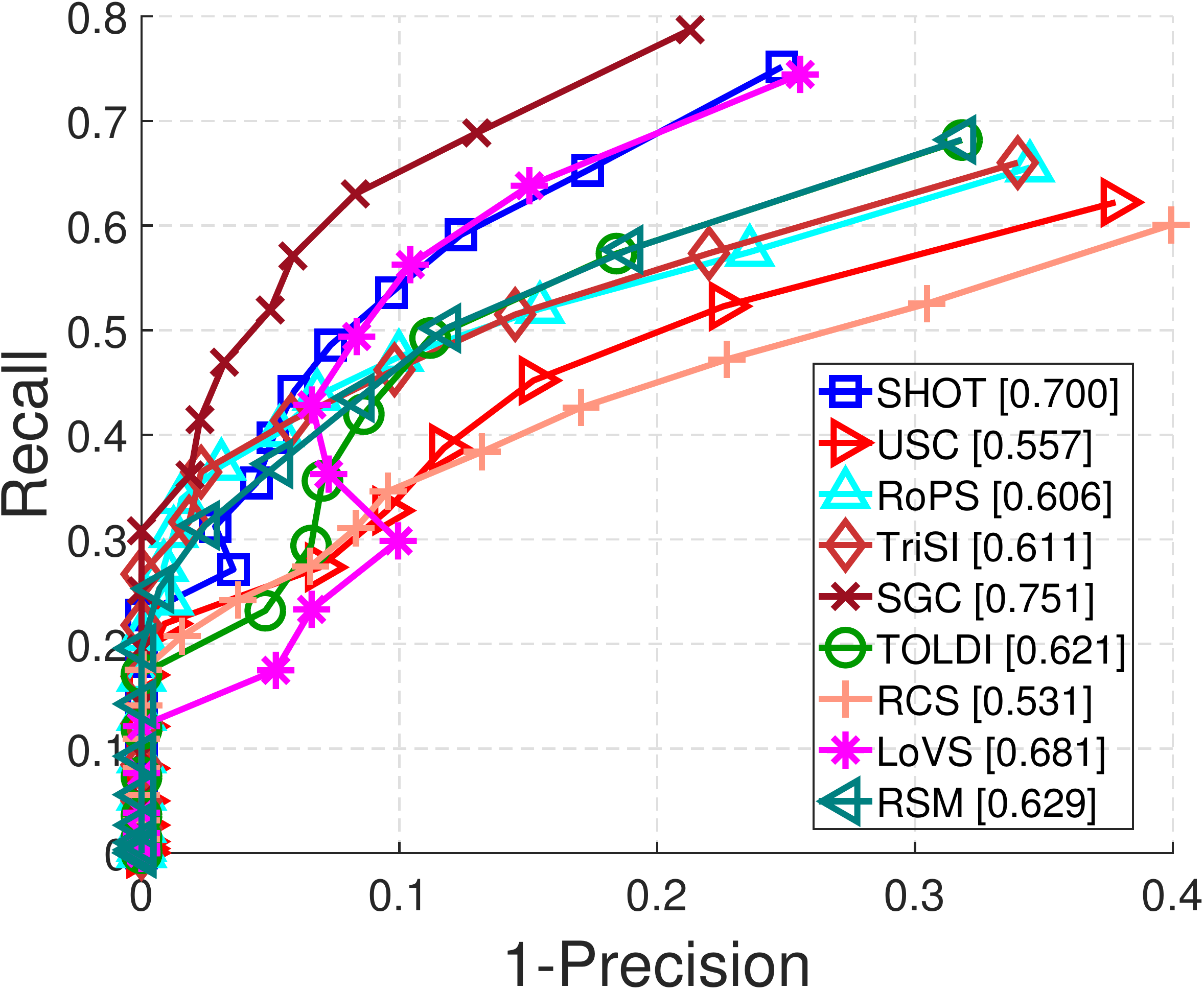}}
	\end{minipage}	
	\begin{minipage}{0.49\linewidth}
		\raggedleft
		\subfigure[\textit{LiDAR Registration}]{
			\includegraphics[width=0.9\linewidth]{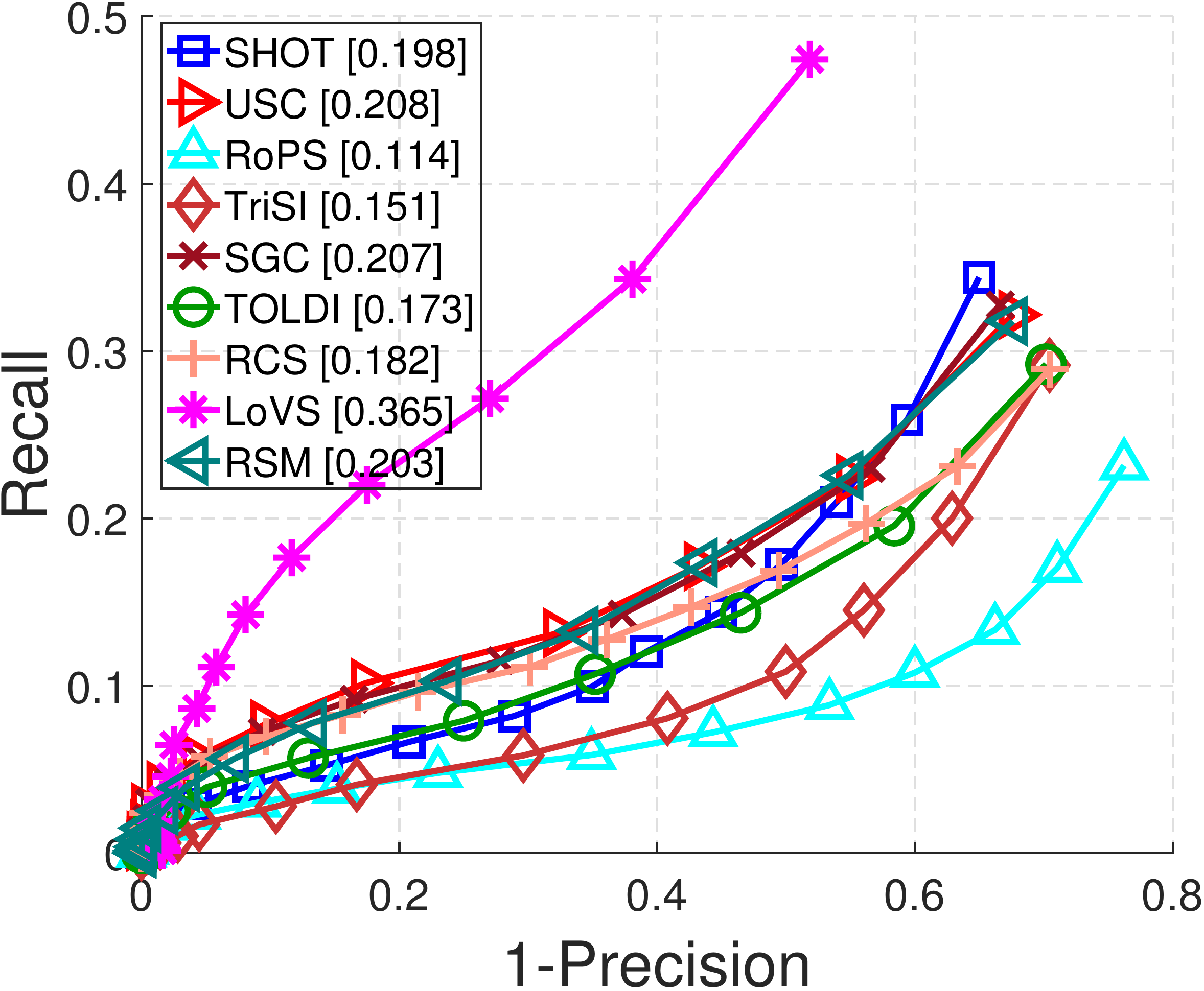}}
	\end{minipage}
	\begin{minipage}{0.49\linewidth}
		\raggedright
		\subfigure[\textit{Kinect Registration}]{
			\includegraphics[width=0.9\linewidth]{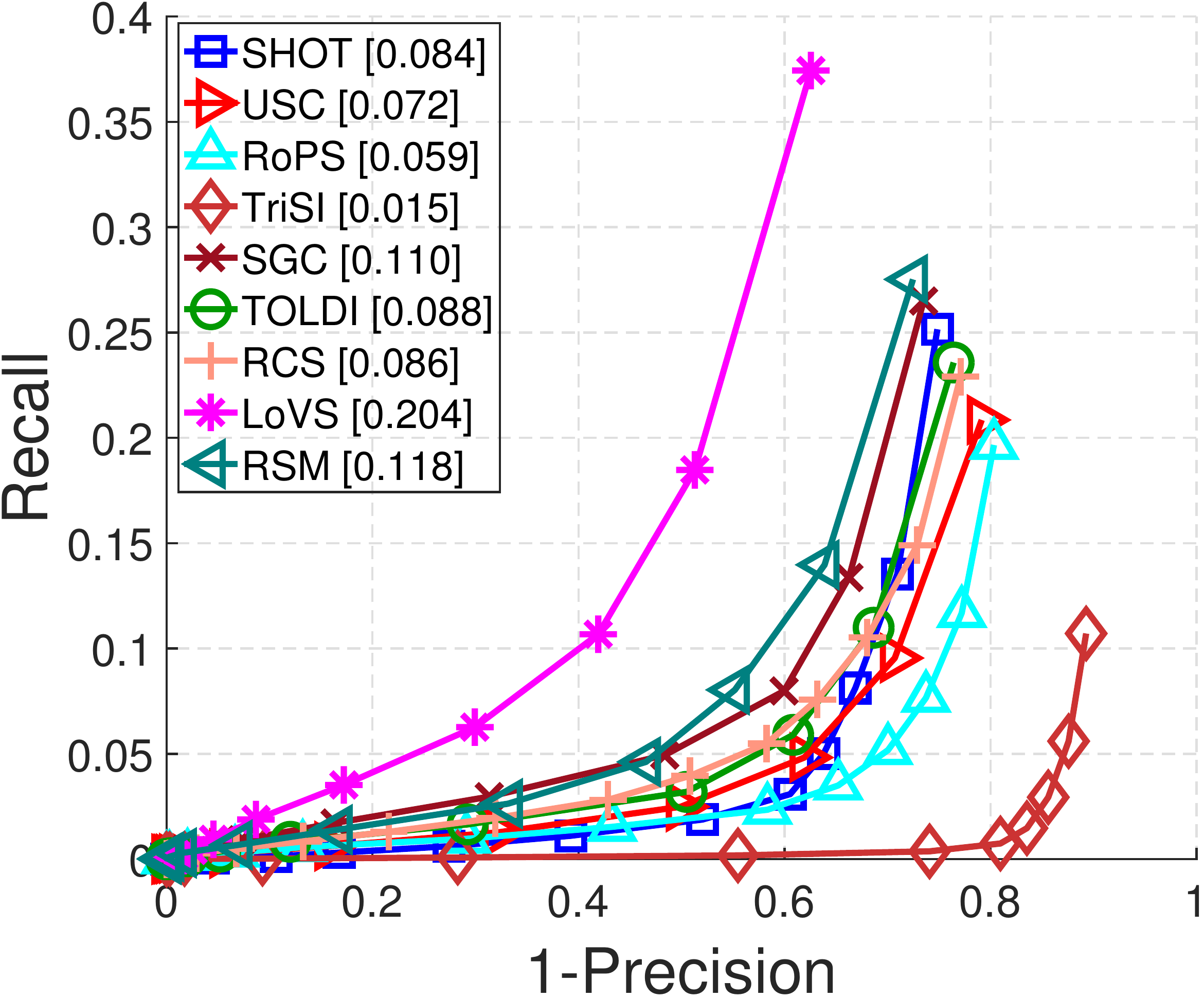}}
	\end{minipage}	
	\hfill
	\caption{Distinctiveness performance of evaluated feature representations on six experimental datasets (numbers in square brackets are AUC values). }
	\label{fig:RPC}
\end{figure}
\subsection{Distinctiveness}
The distinctiveness performance of all evaluated methods on experimental datasets are shown in Fig.~\ref{fig:RPC}. Several observations can be made from the results.

(i) \textit{Retrieval} dataset. The key challenge of this dataset is Gaussian noise. SGC and LoVS are two top-ranked features, marginally surpassing RoPS, RSM, TOLDI, and USC. Obviously, SHOT is significantly inferior to others, because it is composed of attribute histograms from a set of subspaces that may fail to employ the spatial information within  each subspace. As already demonstrated in~\cite{yang2017eval}, spatial information is critical to a feature's distinctiveness. Another interesting finding is that RSM behaves comparable to RoPS (the calculations of these two representations are very similar), though RSM simply relies on silhouette maps rather than density maps. It indicates the information redundancy of RoPS.

(ii) \textit{Laser Scanner} dataset. LoVS neatly surpasses other features on this dataset, followed by SGC. Note that both descriptors perform uniform partition in a cubic volume. LoVS shows better performance than SGC as it solely describes a voxel by judging if there are points inside. By contrast, SGC takes the point count and centroid into consideration that can be easily affected by clutter and occlusion. SHOT and RoPS perform poorly in this context. SHOT lacks enough discriminative power because it uses normal deviation for feature representation that may be ambiguous when ignoring point locations~\cite{yang2017toldi}. The density map encoded by RoPS may change frequently when computed over occluded local surface patches~\cite{Quan2018Representing} due to the significant variation of point count in corresponding patches. In this case, RSM abnegating the point density information for each projected 2D map shows stronger resilience to clutter and occlusion.

(iii) \textit{Kinect} dataset. USC and LoVS  achieve the best and the second best performance, respectively. A common trait of both methods is performing feature description directly in the 3D space instead of performing 3D-to-2D projection, as done in RoPS, TriSI, RSM, and RCS. It is reasonable because projection, though enabling image-like feature representations for 3D point clouds, yet will result in  certain information loss~\cite{Quan2018Local}. Although USC directly performs feature characterization in the 3D space, it relies on point density information which is questionable for this dataset due to the irregular distribution of points. This also results in the performance deterioration of TriSI. Since RoPS also employs point density information but it achieves clear better performance than TriSI, we can infer that further abstracting statistics information (e.g., entropy and moments) from raw density maps produces more robust representations than directly using the density maps.

(iv) \textit{Space Time} dataset. The ranking of evaluated features on this dataset is very similar to that on the \textit{Kinect} dataset. An interesting phenomenon is the behavior of SHOT. Unlike the performance on the \textit{Retrieval} and \textit{Laser Scanner} datasets, it achieves the second best performance datasets on this dataset. It suggests that for point clouds acquired by low-cost sensors that suffer from sparsity, holes, and irregular distributions of points, encoding surface normals is a preferable option. RCS exhibits very limited distinctiveness, indicating that the contour cue turns to be unreliable for such data modality.

(v) \textit{LiDAR Registration} dataset. Clearly, one can see that LoVS outperforms other features by a large margin. LoVS  approximates the spatial structure of with voxel labels, ignoring the information within each voxel. This is shown to be quite robust to partial overlap because the shape geometry inside voxels is vulnerable when local surface patches are incomplete. It has also been demonstrated by SGC, a descriptor that leverages voxel shape information but returns worse performance. The behaviors of other feature representations are similar to each other, except for RoPS.  Owing to limited overlap, the point counts of two corresponding local surfaces may vary significantly, causing a negative impact on the grid values of 2D density maps as employed by RoPS. 

(vi)  \textit{Kinect Registration} dataset. Similar observations on this dataset to those on the \textit{LiDAR Registration} dataset can be made. Specifically, LoVS achieves the best performance, while features based on 2D density maps such as RoPS and TriSI, are inferior to others. However, 2D depth maps, as employed by TOLDI, are revealed to be more robust than 2D density maps.
\subsection{Robustness}
\begin{figure}[t]
	\begin{minipage}{0.49\linewidth}
		\raggedleft
		\subfigure[Varying support radii]{
			\includegraphics[width=0.9\linewidth]{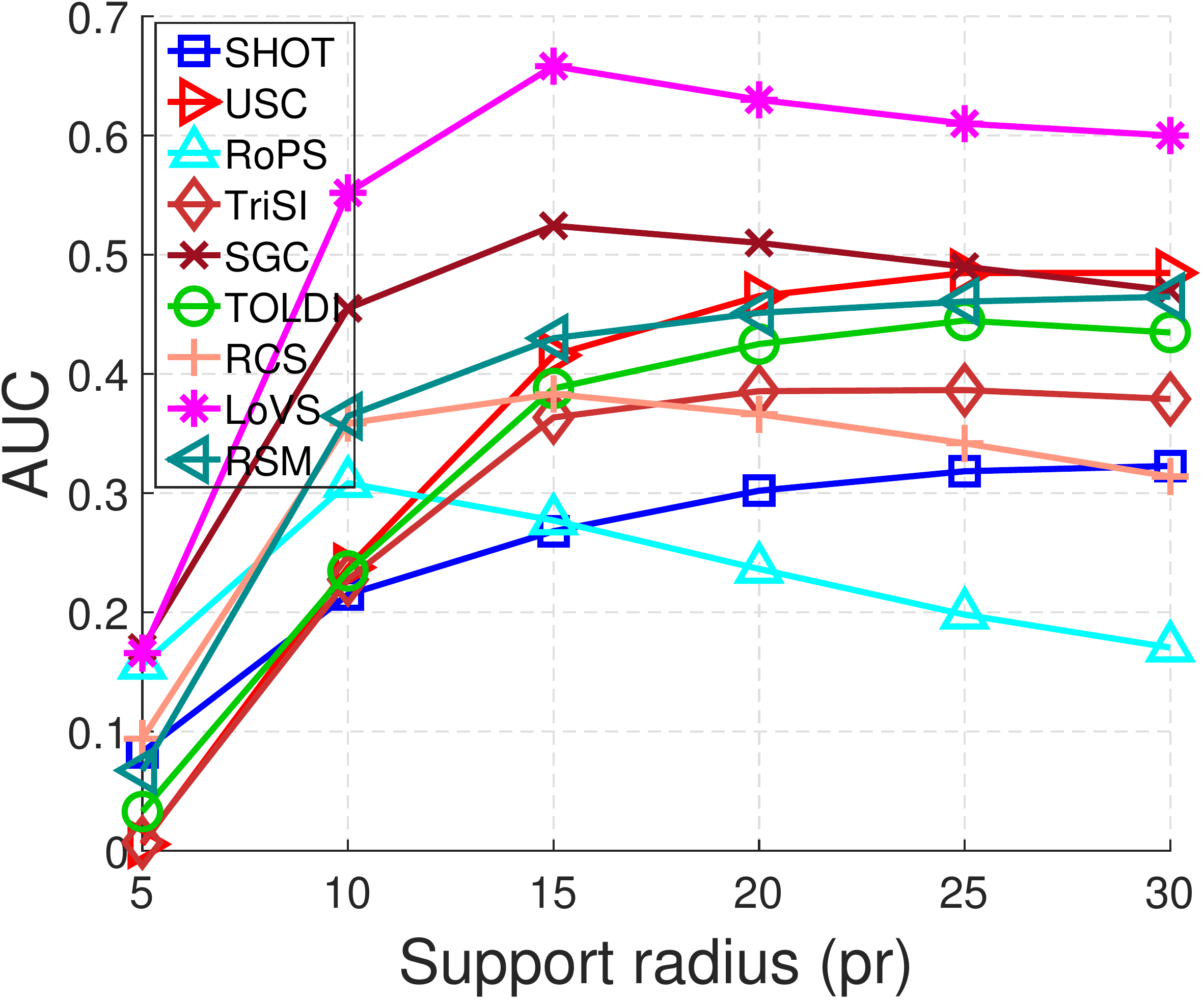}}
	\end{minipage}
	\begin{minipage}{0.49\linewidth}
		\raggedright
		\subfigure[Distance to boundary]{
			\includegraphics[width=0.9\linewidth]{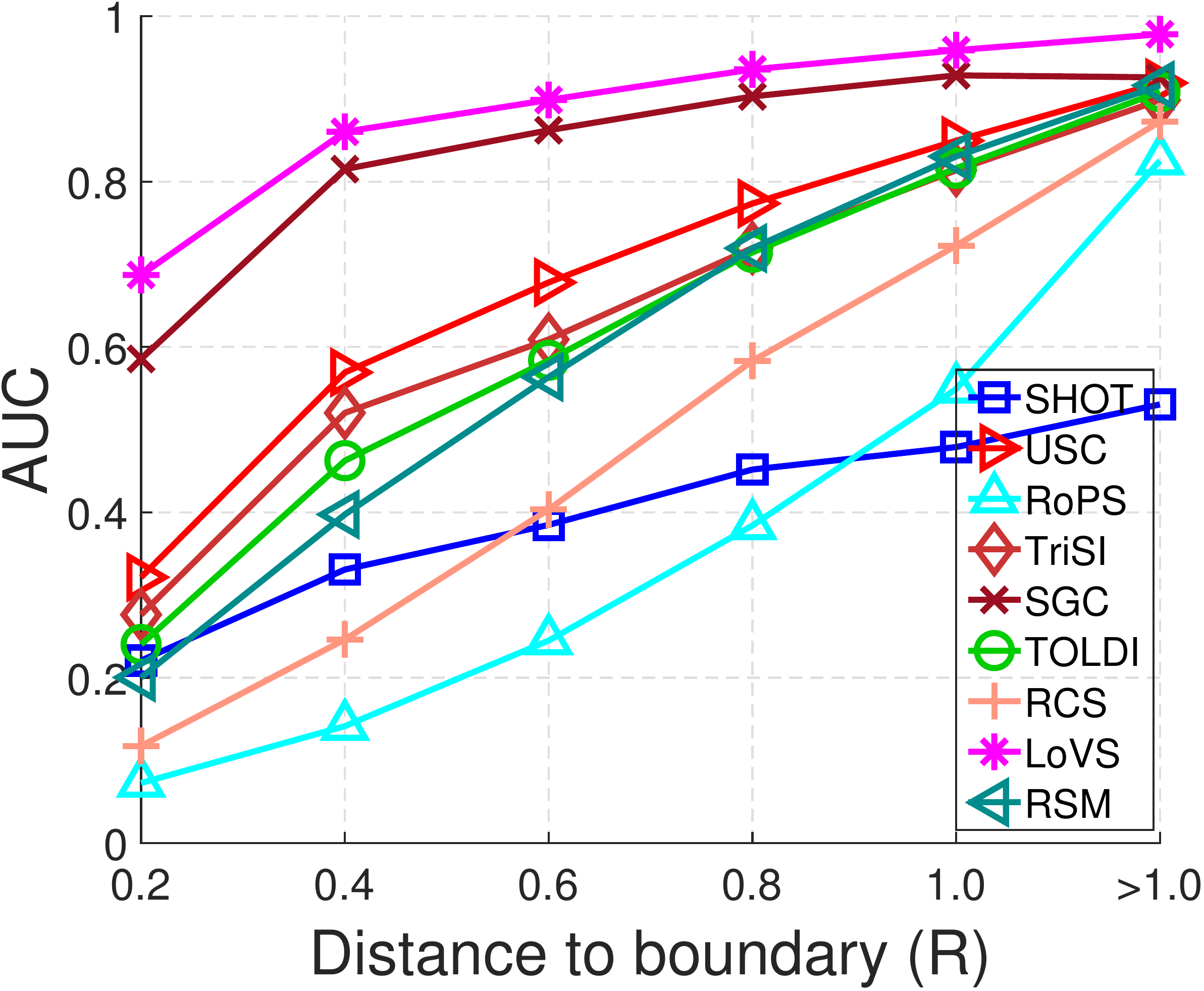}}
	\end{minipage}	
	\begin{minipage}{0.49\linewidth}
		\raggedleft
		\subfigure[Clutter]{
			\includegraphics[width=0.9\linewidth]{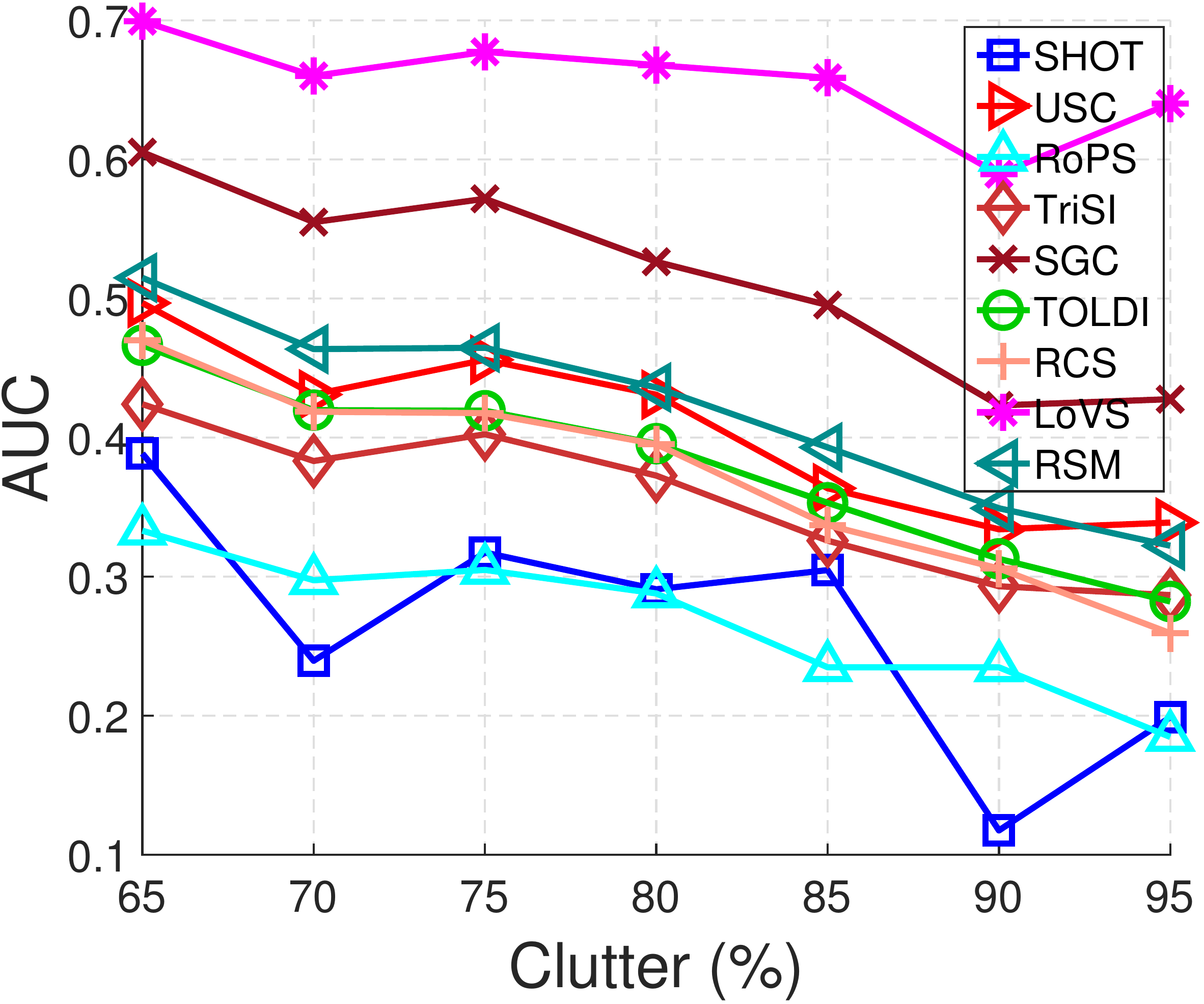}}
	\end{minipage}
	\begin{minipage}{0.49\linewidth}
		\raggedright
		\subfigure[Occlusion]{
			\includegraphics[width=0.9\linewidth]{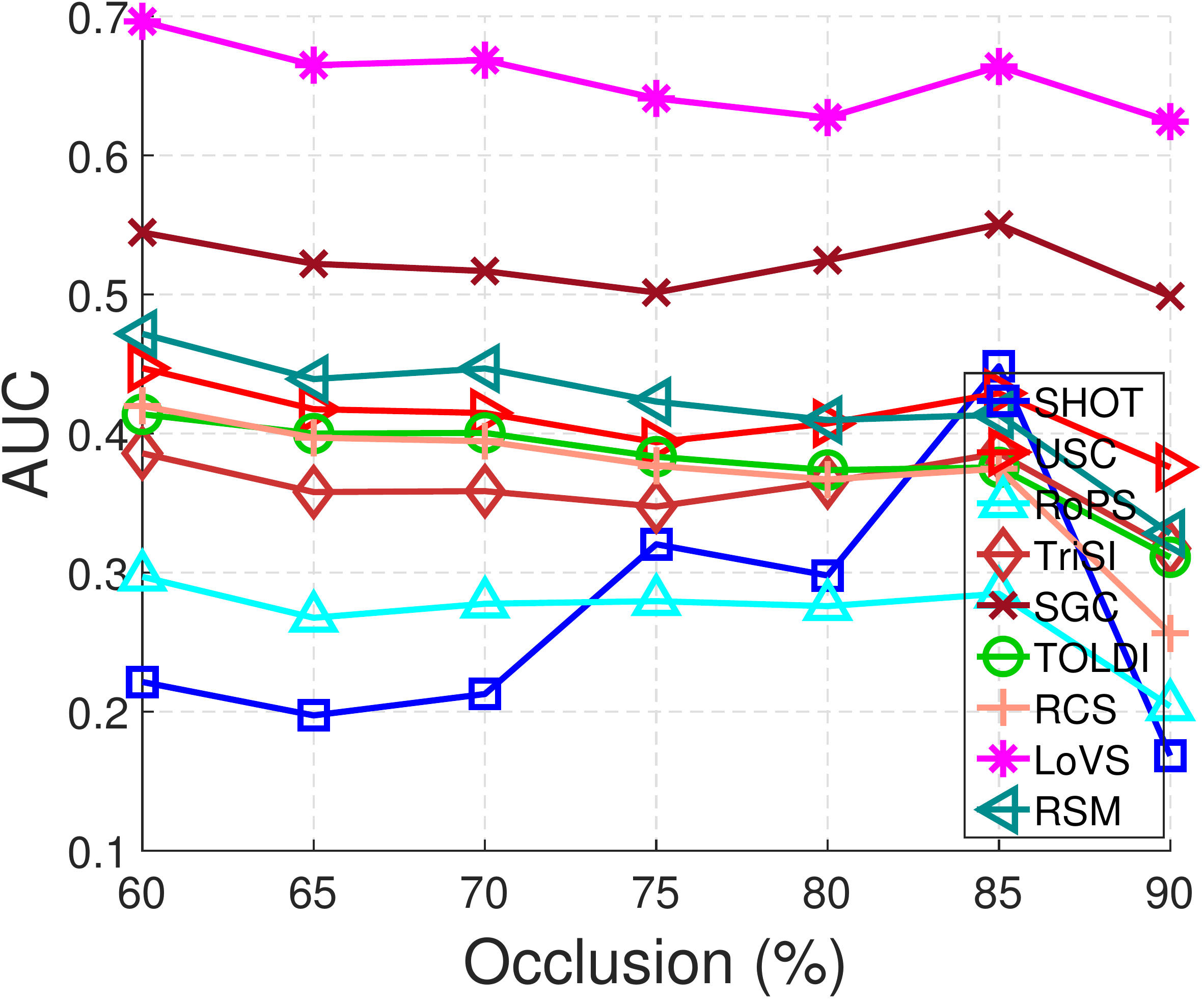}}
	\end{minipage}	
	\begin{minipage}{0.49\linewidth}
		\raggedleft
		\subfigure[Partial overlap]{
			\includegraphics[width=0.9\linewidth]{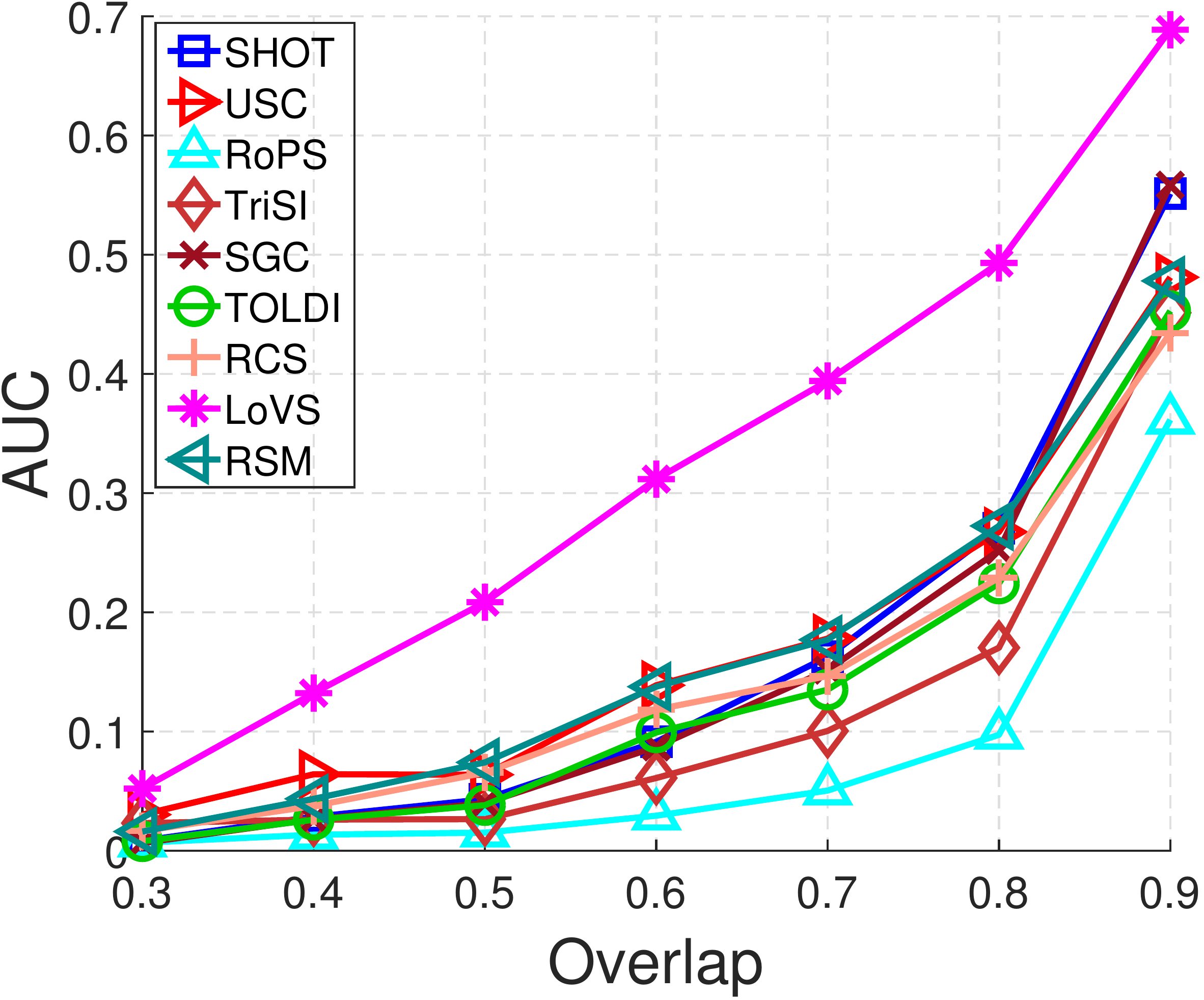}}
	\end{minipage}
	\hfill
	\caption{Robustness performance of evaluated feature representations with respect to \textit{application-dependent} nuisances.}
	\label{fig:robust_c}
\end{figure}
Fig.~\ref{fig:robust_c} reports the results of evaluated descriptors in terms of robustness to application-dependent nuisances. The following phenomenons can be found from the figure.
\begin{figure*}[t]
	\begin{minipage}{0.245\linewidth}
		\raggedleft
		\subfigure[Gaussian noise]{
			\includegraphics[width=0.98\linewidth]{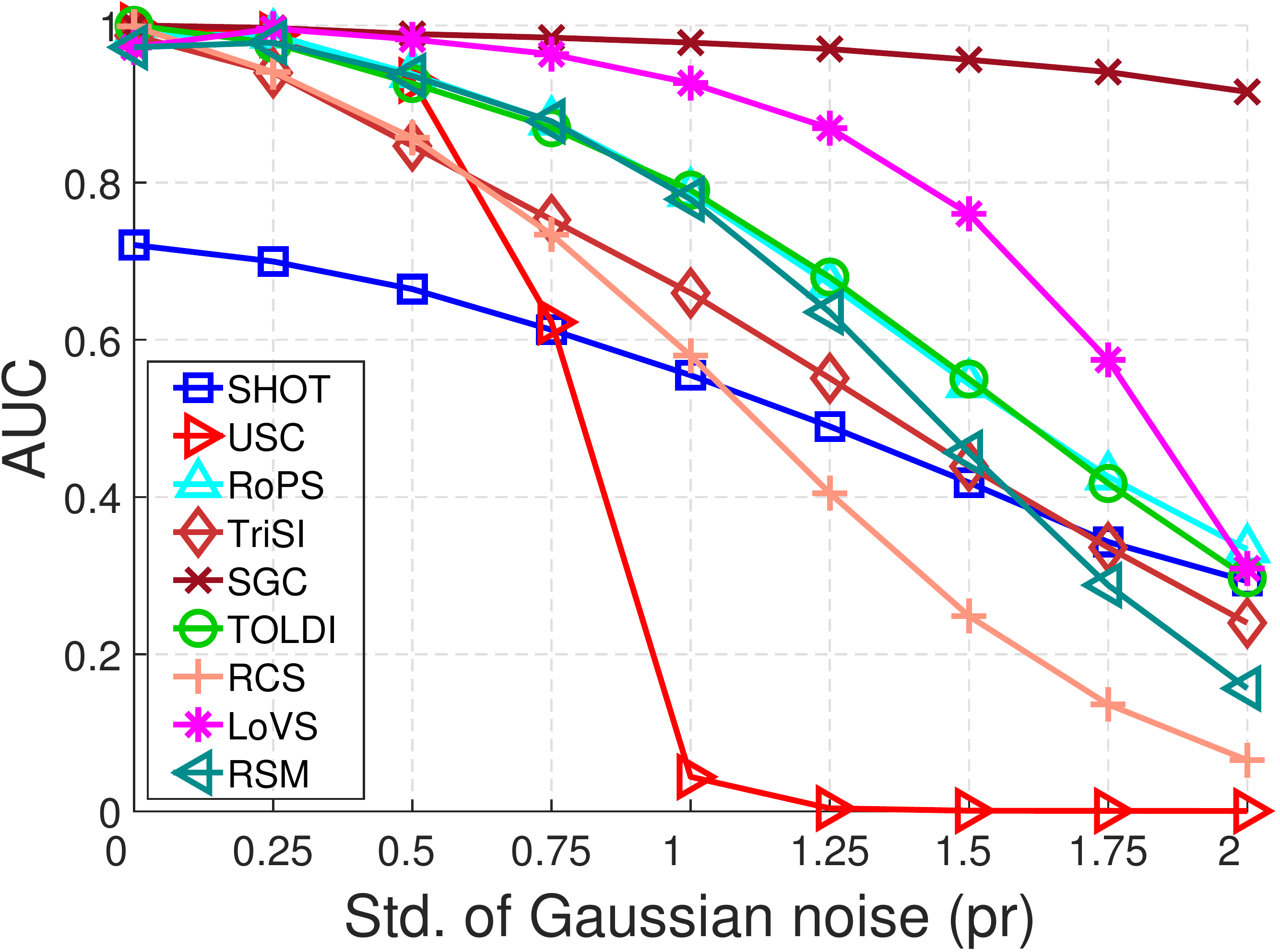}}
	\end{minipage}
	\begin{minipage}{0.245\linewidth}
		\raggedright
		\subfigure[Shot noise]{
			\includegraphics[width=0.98\linewidth]{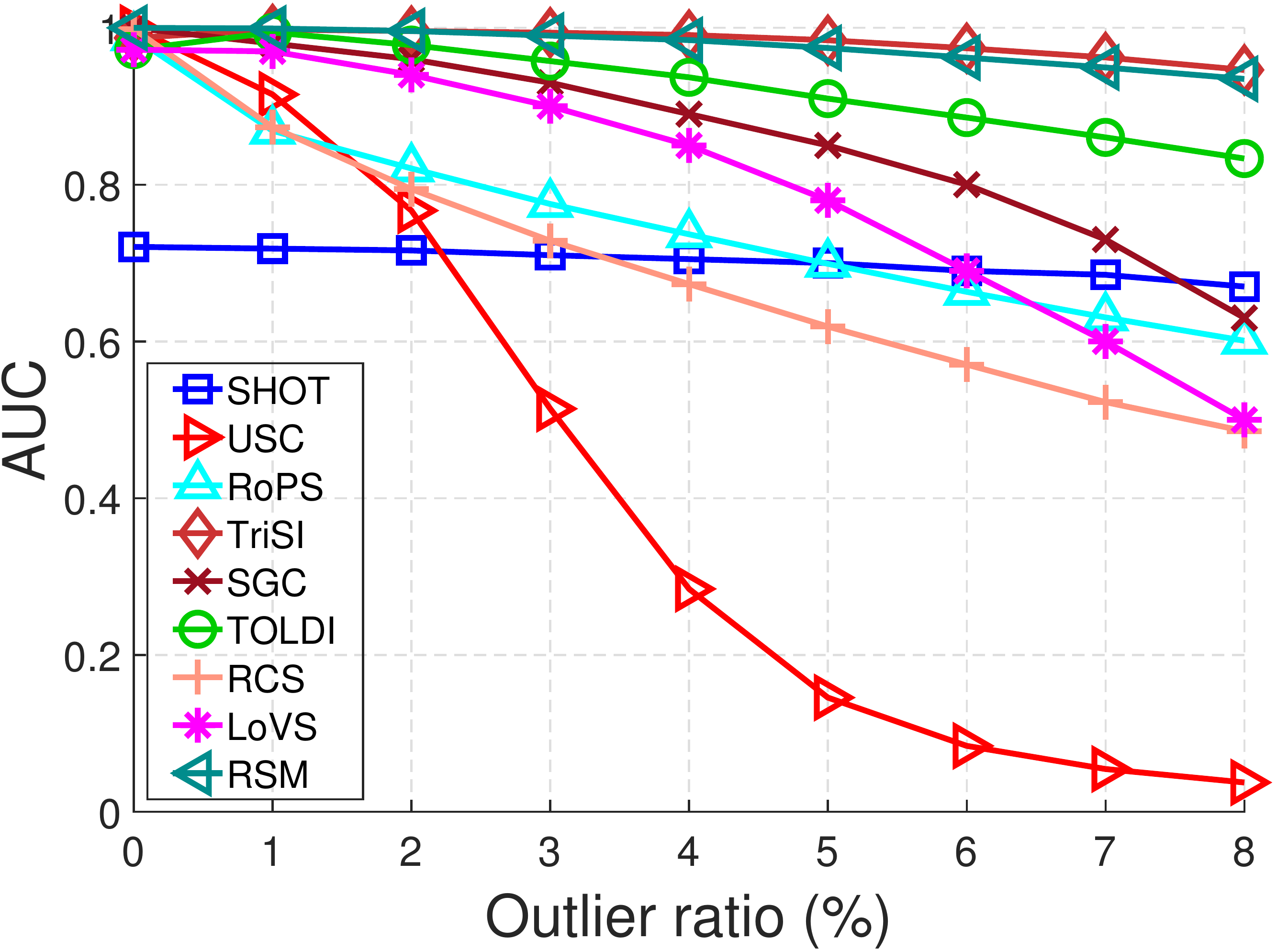}}
	\end{minipage}
	\begin{minipage}{0.245\linewidth}
		\raggedleft
		\subfigure[Uniform data decimation]{
			\includegraphics[width=0.98\linewidth]{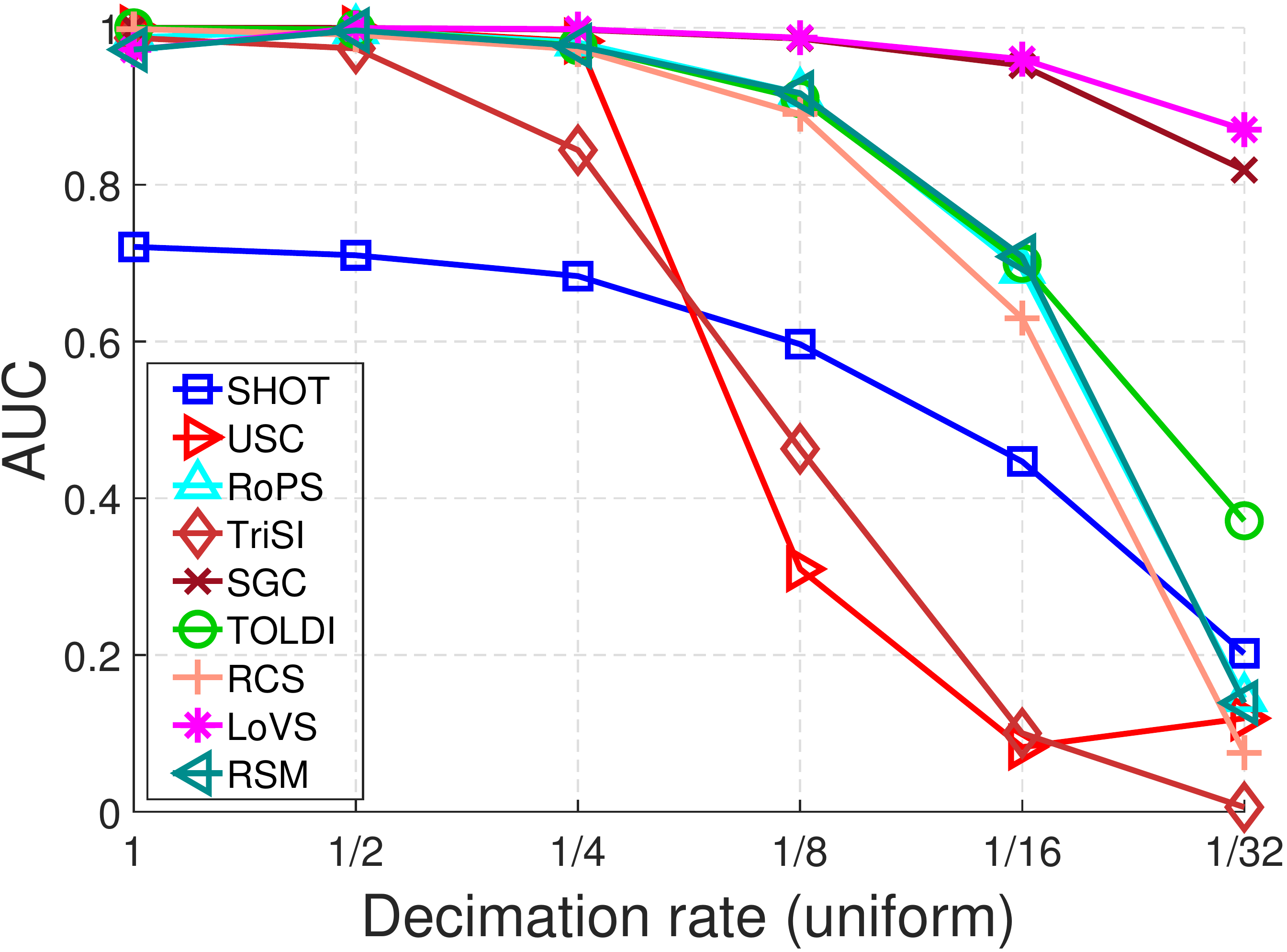}}
	\end{minipage}
	\begin{minipage}{0.245\linewidth}
		\raggedright
		\subfigure[Random data decimation]{
			\includegraphics[width=0.98\linewidth]{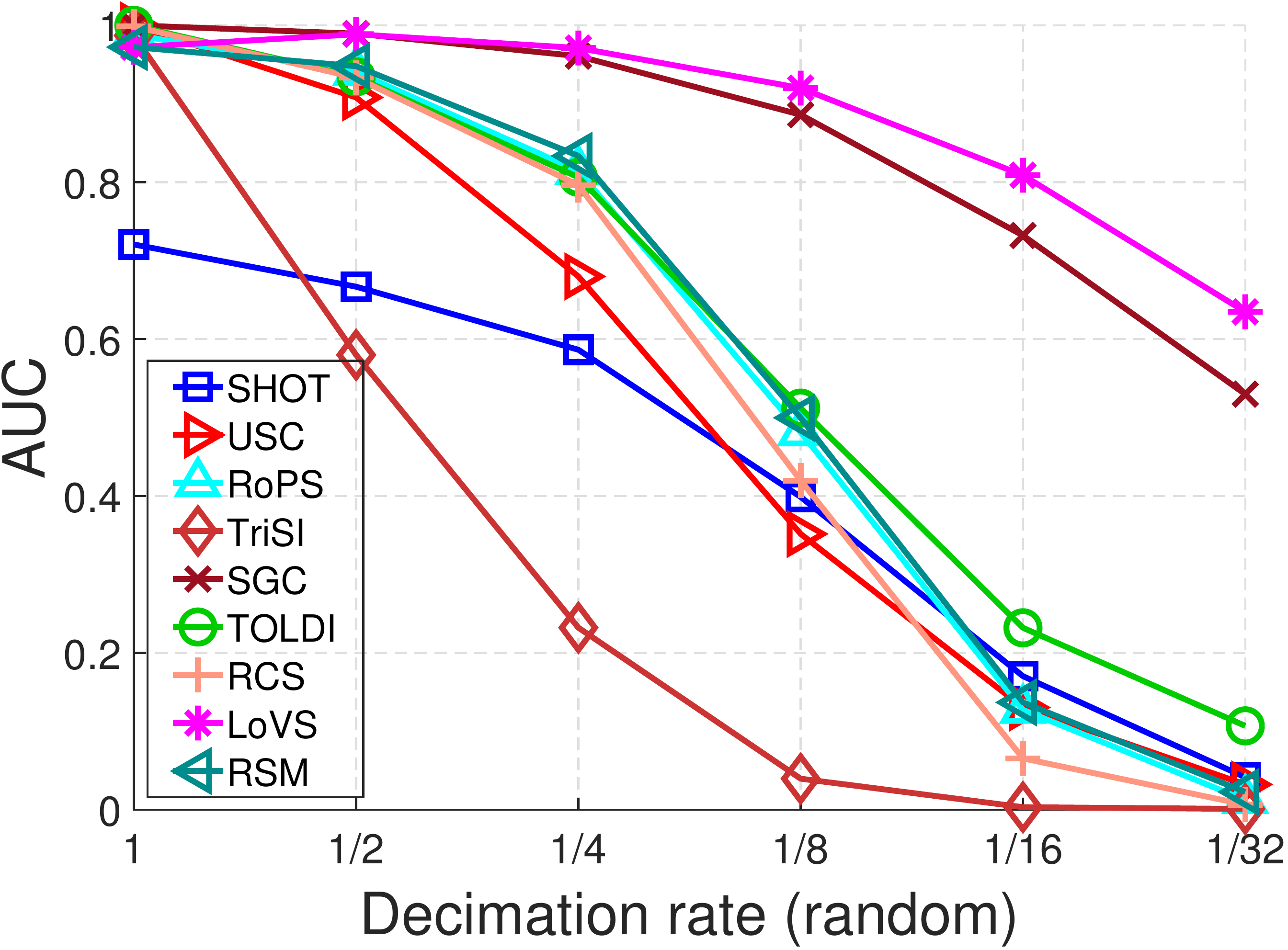}}
	\end{minipage}
	\begin{minipage}{0.245\linewidth}
		\raggedleft
		\subfigure[Keypoint localization error]{
			\includegraphics[width=0.98\linewidth]{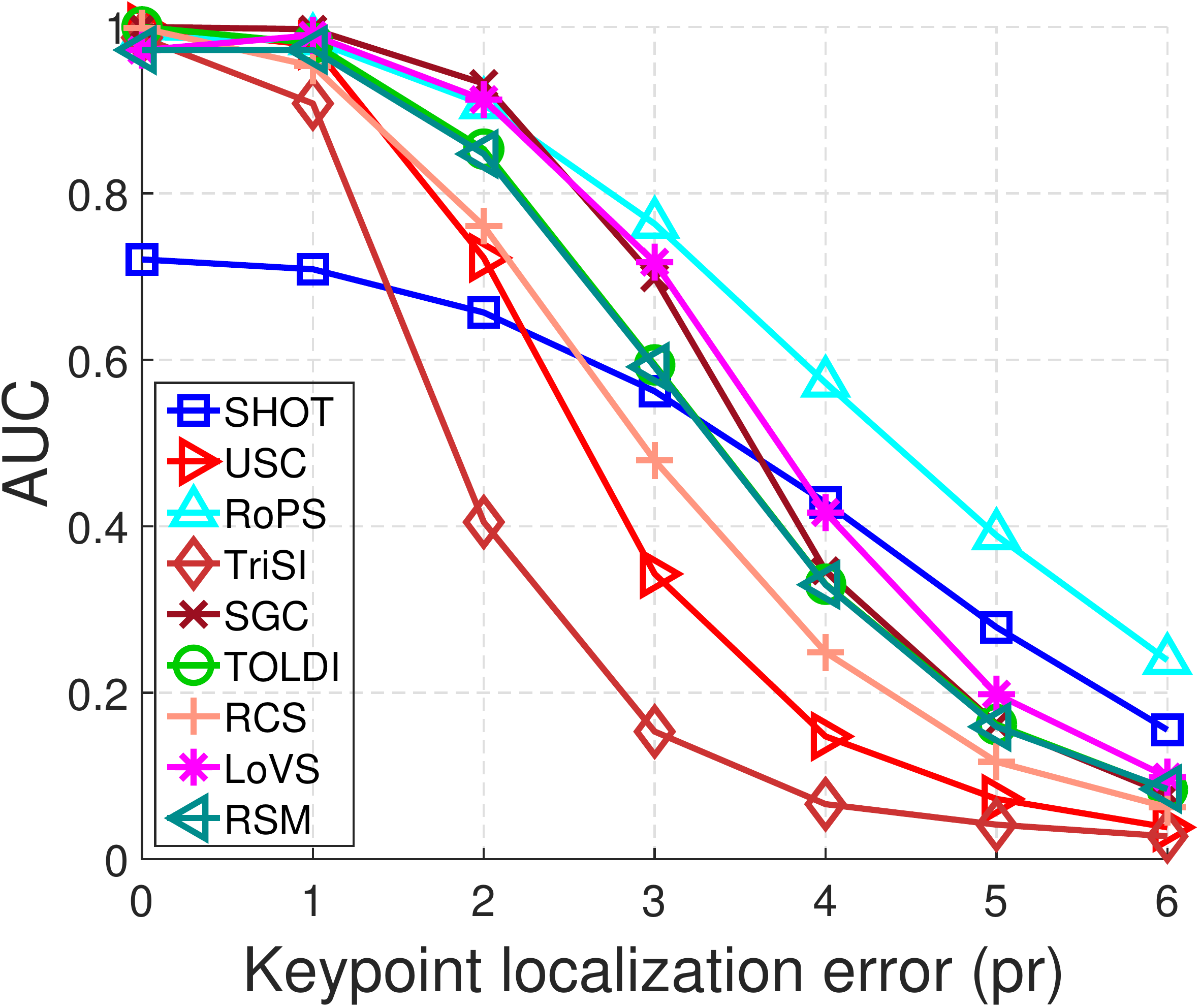}}
	\end{minipage}
	\begin{minipage}{0.245\linewidth}
		\raggedright
		\subfigure[LRF error: $x$-axis]{
			\includegraphics[width=0.98\linewidth]{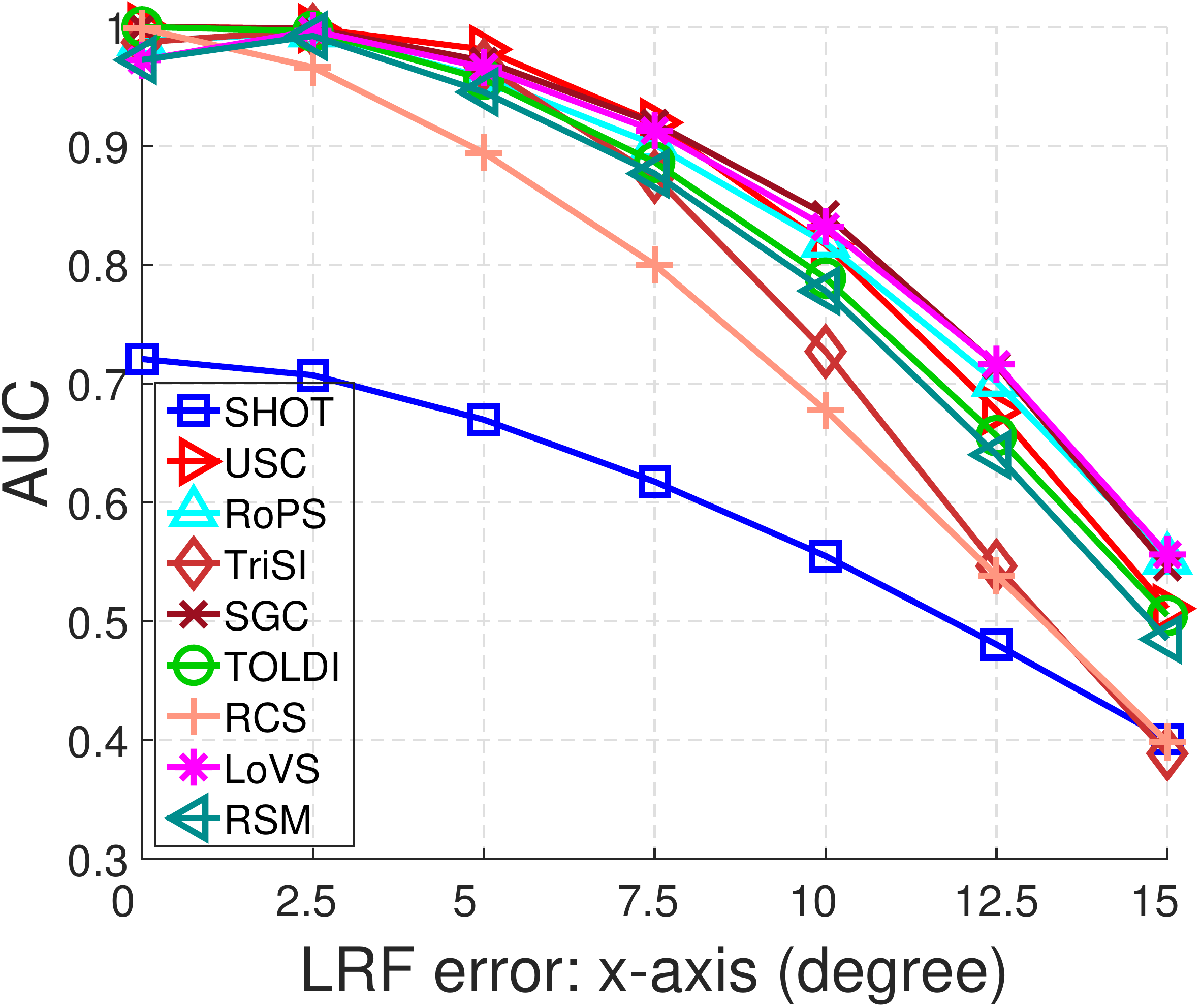}}
	\end{minipage}
	\begin{minipage}{0.245\linewidth}
		\raggedleft
		\subfigure[LRF error: $z$-axis]{
			\includegraphics[width=0.98\linewidth]{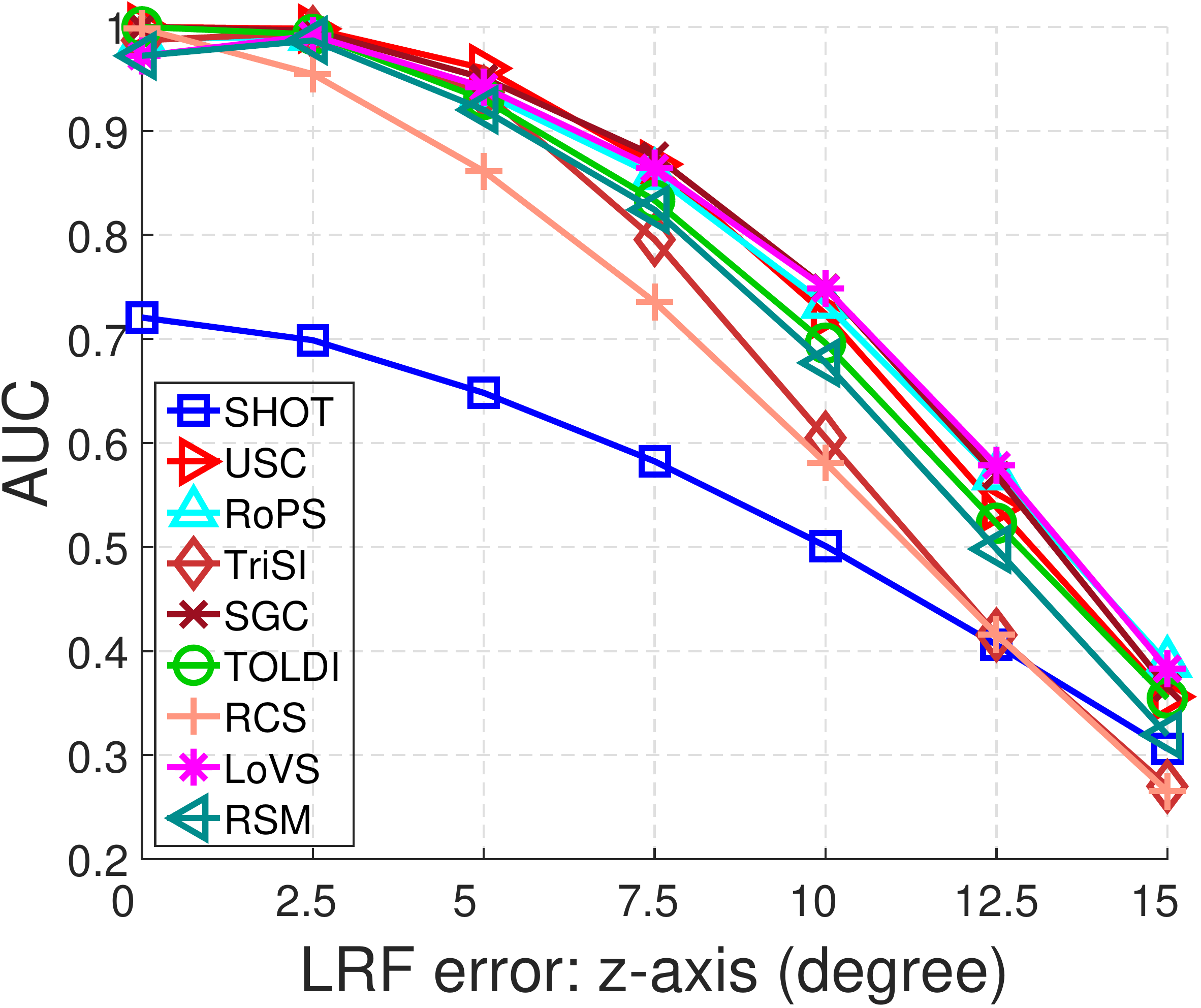}}
	\end{minipage}
	\begin{minipage}{0.245\linewidth}
		\raggedright
		\subfigure[LRF error: $x$\&$z$-axis]{
			\includegraphics[width=0.98\linewidth]{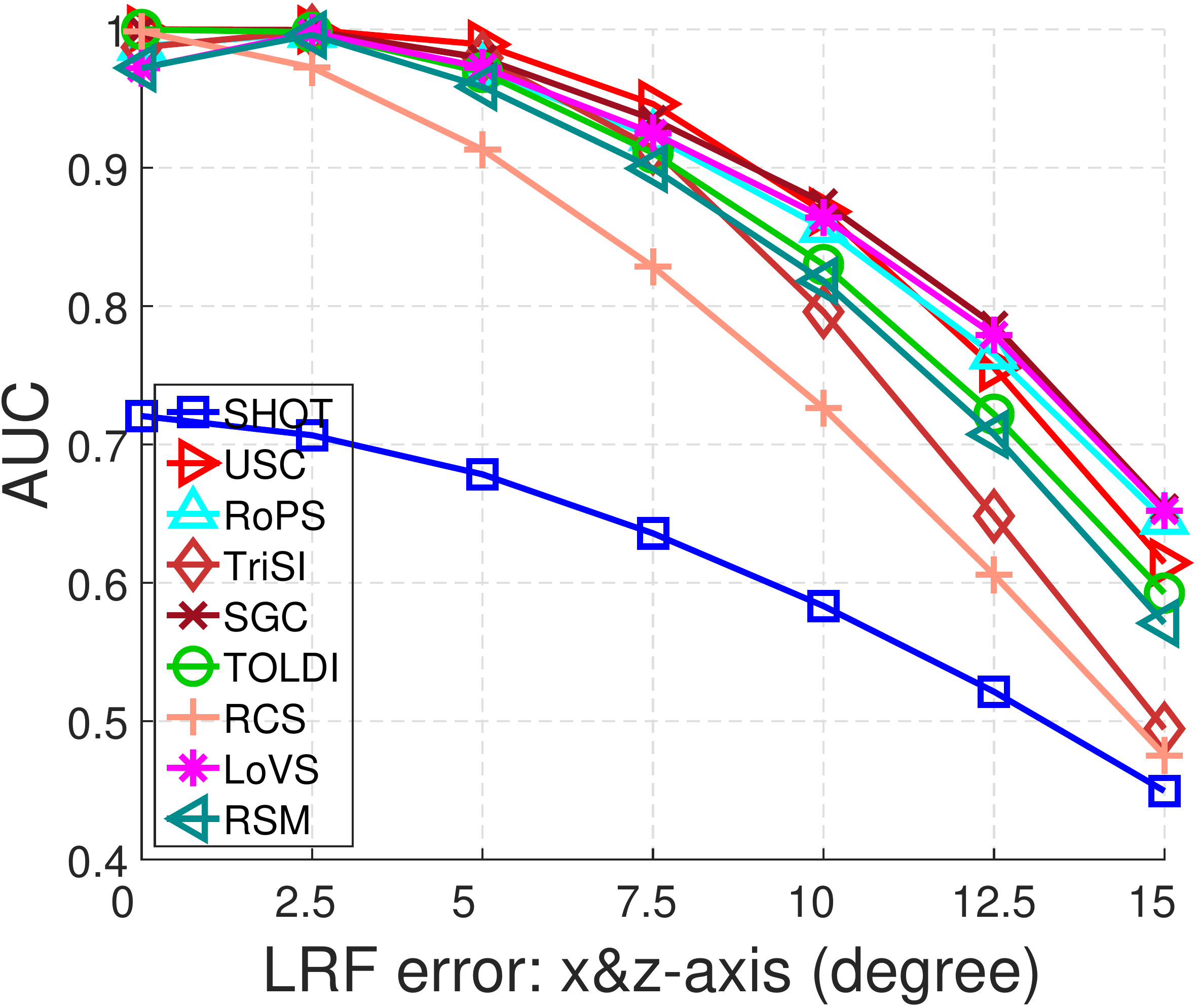}}
	\end{minipage}
	\hfill
	\caption{Robustness performance of evaluated feature representations with respect to \textit{application-independent} nuisances.}
	\label{fig:robust_uc}
\end{figure*}

(i) \textit{Varying support radii}. When the support radius $R$ is 5 \textit{pr}, LoVS, SGC, and RoPS achieve better performance than others. USC and TriSI almost fail to retrieve any correct matches at this scale. As $R$ further increases, LoVS consistently surpasses other competitors. The performance of RoPS drops quickly as $R$ is greater than 10 \textit{pr}, while the variation of performance for other features is mild. A general trend for all features is that their performance first monotonically increases with $R$ as more geometric information can be encoded, but then degrades when $R$ further gets larger because of the perturbation caused by clutter. Overall, LoVS and SGC are the two most robust features to the variation of support radius.

(ii) \textit{Distance to boundary}. The more closer to the boundary, the more challenging the keypoints are for feature matching. Local surfaces in boundary regions usually suffer from data missing, which well explains the limited performance by RoPS and TriSI because 2D density maps are not consistent for corresponding local surface patches. In this case, LoVS and SGC clearly outperform all others. For keypoints  at a distance larger than the support radius $R$ to the boundary, i.e., corresponding surface patches are relatively complete, all methods except for SHOT behave comparable with each other. Nonetheless, the performance variation of SHOT as the distance to boundary changes is less significant than most features, arising from the fact that normal deviation histograms from non-empty subspaces  remain stable.

(iii) \textit{Clutter and occlusion}. In terms of the robustness to clutter and occlusion, it is noteworthy that although RoPS and TriSI are particularly designed to resist the impact of clutter and occlusion, they cannot handle these nuisances effectively. Although RoPS and TriSI (the overall descriptors) are demonstrated to be very effective for 3D object recognition~\cite{guo2013rotational,guo2015novel}, their feature representations are obviously not prior options. Interestingly, LoVS is originally designed for the pairwise registration of point clouds, though, it is demonstrated to be the best option as well for 3D object recognition in cluttered and occluded scenes. Our results may motivate the researchers to rethink the traditional approaches or rules for promoting feature representations' robustness to some particular nuisances. 

(iv) \textit{Partial overlap}. For data pairs with very limited overlap, i.e., an overlap ratio of 0.3, all features exhibit quite poor performance. In less challenging cases with higher overlap ratios, LoVS becomes the best performed feature representation and exceeds others by a significant gap. RoPS and TriSI are the two most inferior features. We  can therefore conclude that when the local surface undergoes data missing (mainly caused by limited overlap and occlusion) or data expansion (mainly caused by clutter), 2D density maps are very susceptible meta-representations.

Regarding the evaluation of the robustness to application-independent nuisances, as shown in Fig.~\ref{fig:robust_uc}, we can make the following observations.

(i) \textit{Gaussian noise}. When the standard deviation of Gaussian noise is smaller than 0.75 \textit{pr}, SGC and LoVS are two outstanding features as their AUC values remain almost unaltered. As the Gaussian noise becomes more severe, the SGC feature achieves the best performance and the margin between SGC and the second best one is particularly large when the standard deviation is greater than 1.5 \textit{pr}, while LoVS meets a clear performance degradation. It is owing to the robustness of geometric centroids to noise~\cite{Tang2016Signature}. However, noise with large standard deviations can frequently change the voxel labels of LoVS because noisy points may exist in originally empty voxels. The performance of USC drops rapidly as well with severe noise. This is because USC splits the local volume in a more fine-grained manner than SGC and LoVS, leading to an increased sensitivity to noise. Although normals are demonstrated to be sensitive to noise~\cite{yang2016fast,guo2013rotational}, SHOT encoding normal deviation information is even more stable than several competitors such as TriSI, RSM, and RCS as the standard deviation of noise exceeds 1.5 \textit{pr}. 

(ii) \textit{Shot noise}. TriSI, RSM, and SHOT show very stable performance for data with outliers. TriSI is based on 2D point density maps and outliers hold low opportunities to significantly change the density of a grid; RSM alleviates the influence of outliers via 3D-to-2D projection and the exclusion of isolated 1-labeled pixels in each silhouette image; SHOT relies on surface normals that are computed by performing PCA on a subset of nearest neighbors, whereas outliers are distant points from the 3D surface and not likely to be included for normal calculation. We can see that RoPS is inferior to TriSI although both features resort to 2D density maps, the difference is that RoPS further extracts statistics information for these density maps, shortening the feature length yet sacrificing the robustness to outliers. Another finding is the susceptibility of signature-based features with 3D spatial information to shot noise, i.e., SGC, LoVS, and USC. Common to all the three features is the fact that a single point in a 3D subspace may significantly affect the bin value of the feature.

(iii) \textit{Uniform data decimation}. When reducing the data volume to $\frac{1}{2}$ and $\frac{1}{4}$ of its original volume, the performance of most feature representations except for TriSI keeps very stable. As more points are uniformly extracted, LoVS achieves the best performance, followed by SGC and TOLDI; USC and TriSI exhibit limited performance in such case. Besides USC, all features have designed particular rules to achieve robustness to data decimation. For instance, TriSI and RoPS have performed normalization on the 2D density maps; SHOT employs statistical histograms; LoVS, TOLDI, RSM, and RCS only consider one representative point for bin value assignment. Our results indicate that among all these rules, the solutions proposed by LoVS and SGC are more effective.

(iv) \textit{Random data decimation}. Compared with uniform data decimation, random data decimation possesses an additional challenge, i.e., non-uniformity. Few feature representations have noticed this nuisance that is common in real-world scanned data. Experimental results show that LoVS is the most robust feature to random data decimation, followed by SGC. This is benefited from the fact that LoVS assigns feature values to voxels based on judging whether a voxel contains points but ignores the exact point count. As a result, neither uniform nor random data decimation is expected to significantly vary the feature value of a voxel. The impact on TriSI is especially obvious, i.e., it drops to the most inferior one with $\frac{1}{2}$ random data decimation. Although it has performed normalization for each 2D density map, but it is based on the assumption that the density value of each grid changes simultaneously with the total number of points and therefore fails to cope with non-uniformity.

(v) \textit{Keypoint localization error}. When the keypoint localization error $d_{key}$ is within 2 \textit{pr}, SGC, LoVS, and RoPS are more robust than others. As $d_{key}$ further increases, RoPS surpasses all others. It is noteworthy that although SHOT is less distinctive than other features, but it achieves the second best performance when $d_{key}$ is greater than 4 \textit{pr}. It is because keypoint localization error will lead to certain differences between corresponding surface patches; signature-based methods, e.g., LoVS, SGC, TOLDI, and RCS, are very sensitive to such changes because their feature values are often determined by some particular points that  are difficult to keep consistent in this context; the effect on features based on statistical histograms, e.g., RoPS and SHOT, are less obvious as the main statistical information between two coarsely corresponding surface patches are supposed to be similar.

(vi) \textit{LRF error}. The ranking of features with respect to LRF errors from the $x$-axis, $z$-axis, and both axes are generally similar. Specifically, RoPS, SGC, and LoVS are more stable than others. For RoPS, a potential explanation  is that it abstracts more robust statistics from initial density features; for SGC and LoVS, they coarsely split the local cubic volume into uniform voxels and slight LRF errors are not likely to significantly change the voxel labels and geometric centroids  employed by the two features. The behaviors of TriSI and RCS are relatively sensitive to LRF errors. Specifically, the performance of RCS drops clearly even with small LRF errors, arising from the fact that RCS characterizes the geometry of each projected 2D map with a set of contour points. Unfortunately, the coordinates of these contour points would change simultaneously with unrepeatable LRFs, resulting a significant variation in the final feature.
\subsection{Compactness}

The compactness results reflected by AUC versus storage of evaluated feature representations are presented in Fig.~\ref{fig:compact}.

Among all evaluated features, LoVS, RSM, RCS, and RoPS  require less storage than others. LoVS and RSM are two binary features, and RCS and RoPS are two compact real-valued features. A light-weight feature is  expected to  achieve a good balance between storage occupancy and distinctiveness. On the \textit{Retrieval} dataset, LoVS, RSM, and RoPS are three superior feature against others. On the \textit{Space Time} dataset, we find that none of tested features manage to strike a well balance between compactness and descriptiveness. On the other four experimental datasets, it is salient that LoVS dramatically surpasses others on these datasets. Although SGC usually shows competitive distinctness to LoVS, the results suggest it has great redundancy. It is interesting to note that SGC additionally considers the geometric center and point count attributes of a voxel, yet resulting less robustness and more storage occupancy. We believe further investigations should be made on the relationship between spatial partition and feature description within a subspace.
\begin{figure}[t]
	\begin{minipage}{0.49\linewidth}
		\centering
		\subfigure[\textit{Retrieval}]{
			\includegraphics[width=\linewidth]{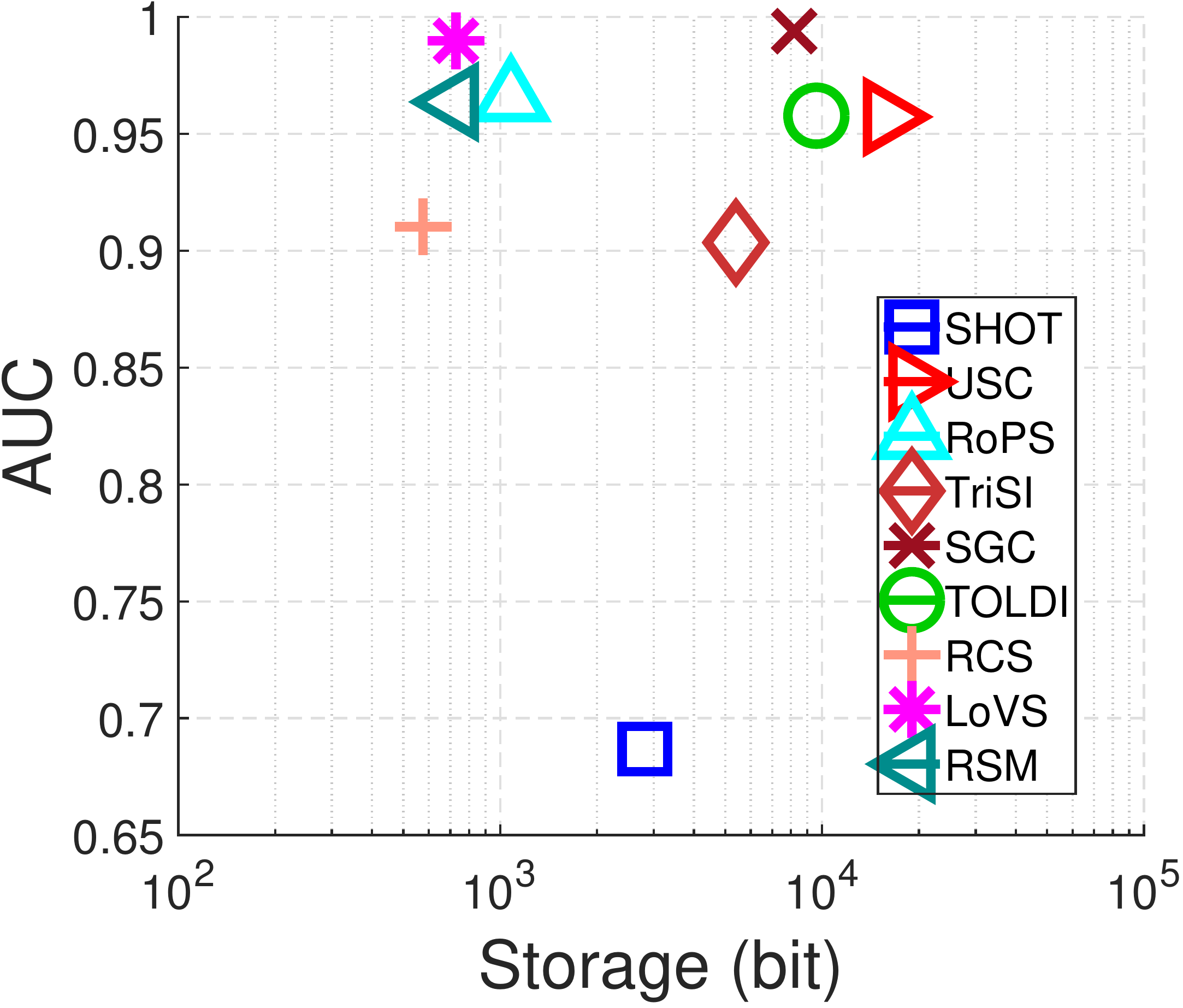}}
	\end{minipage}
	\begin{minipage}{0.49\linewidth}
		\centering
		\subfigure[\textit{Laser Scanner}]{
			\includegraphics[width=\linewidth]{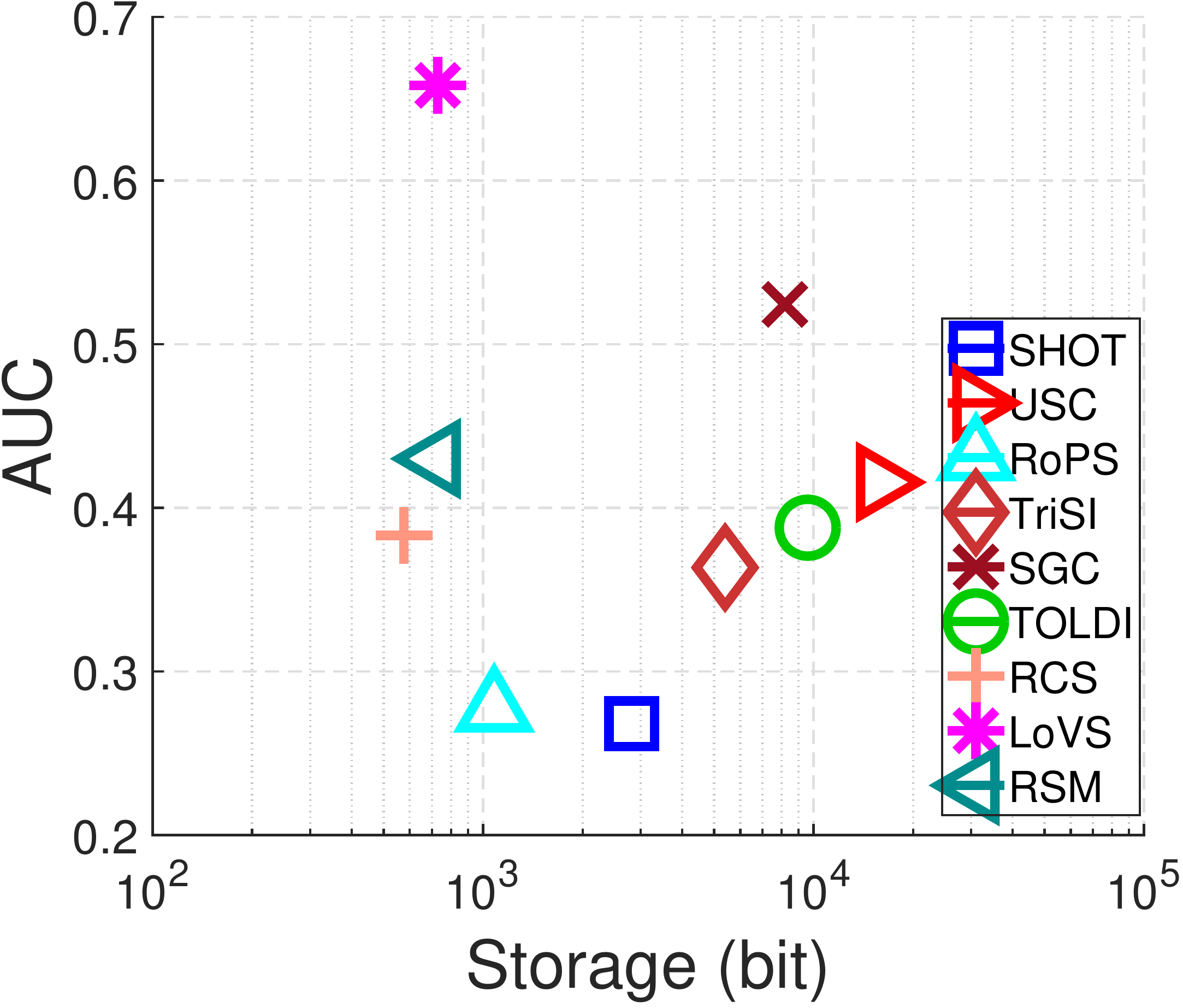}}
	\end{minipage}	
	\begin{minipage}{0.49\linewidth}
		\centering
		\subfigure[\textit{Kinect}]{
			\includegraphics[width=\linewidth]{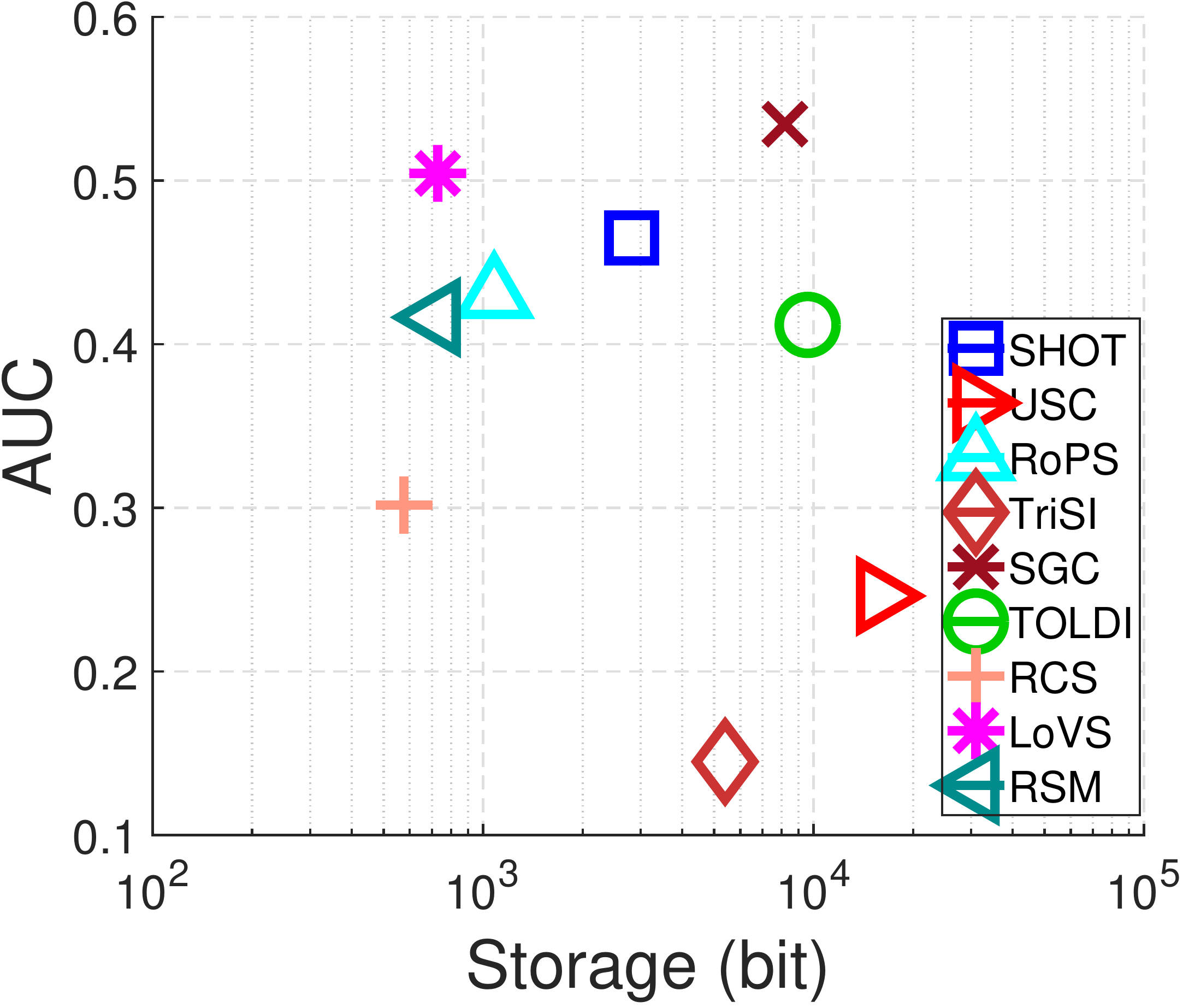}}
	\end{minipage}
	\begin{minipage}{0.49\linewidth}
		\centering
		\subfigure[\textit{Space Time}]{
			\includegraphics[width=\linewidth]{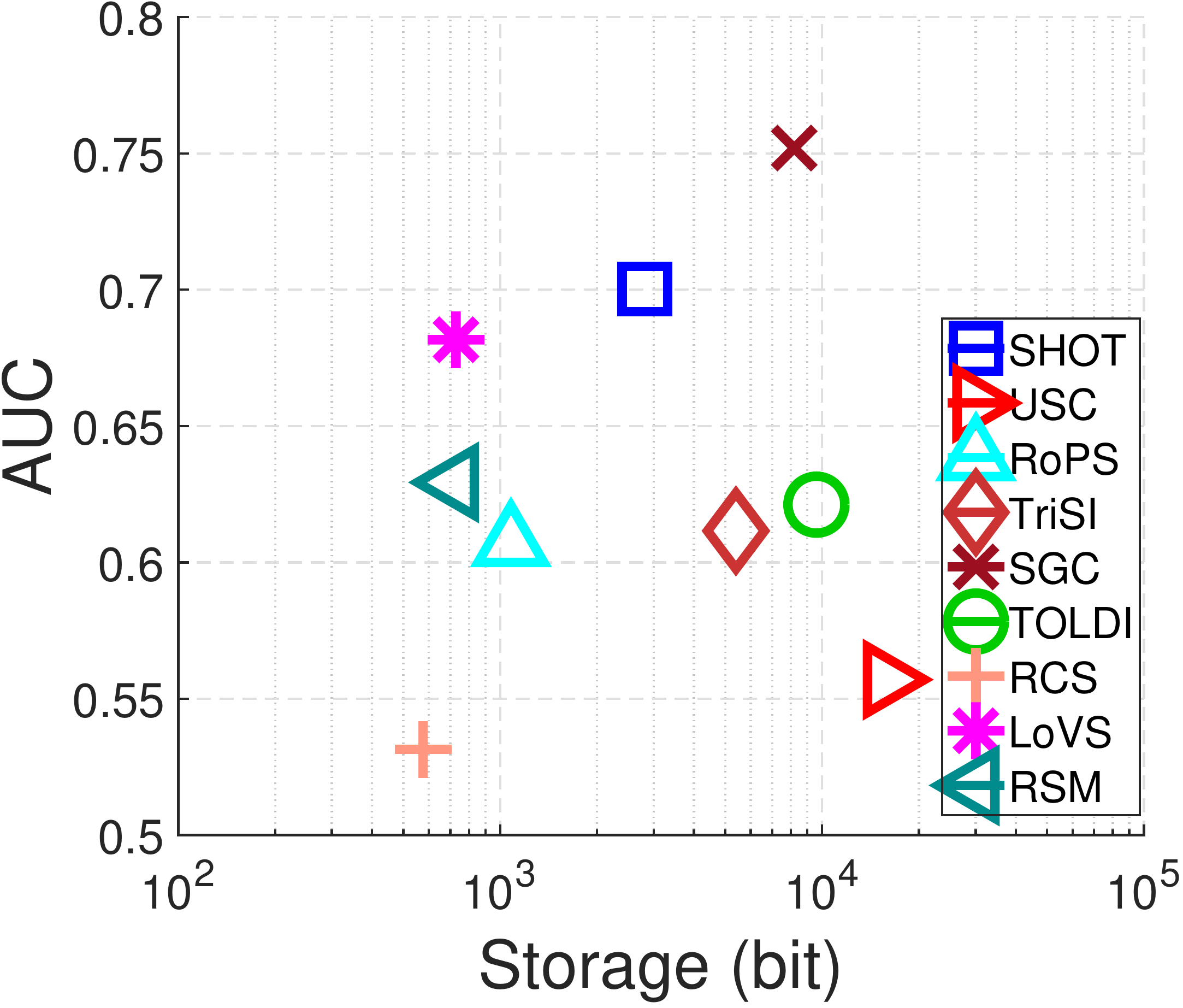}}
	\end{minipage}	
	\begin{minipage}{0.49\linewidth}
		\centering
		\subfigure[\textit{LiDAR Registration}]{
			\includegraphics[width=\linewidth]{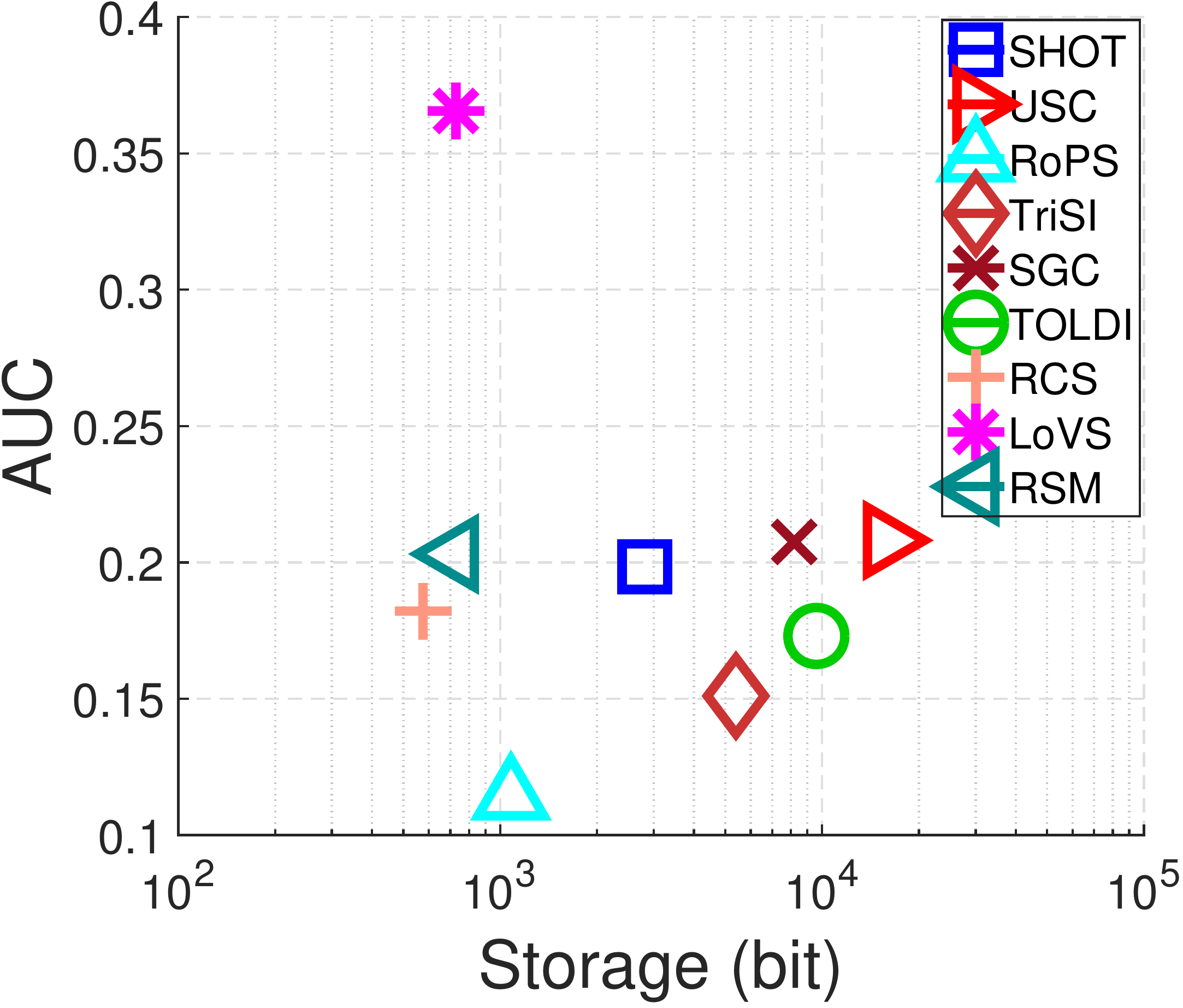}}
	\end{minipage}
	\begin{minipage}{0.49\linewidth}
		\centering
		\subfigure[\textit{Kinect Registration}]{
			\includegraphics[width=\linewidth]{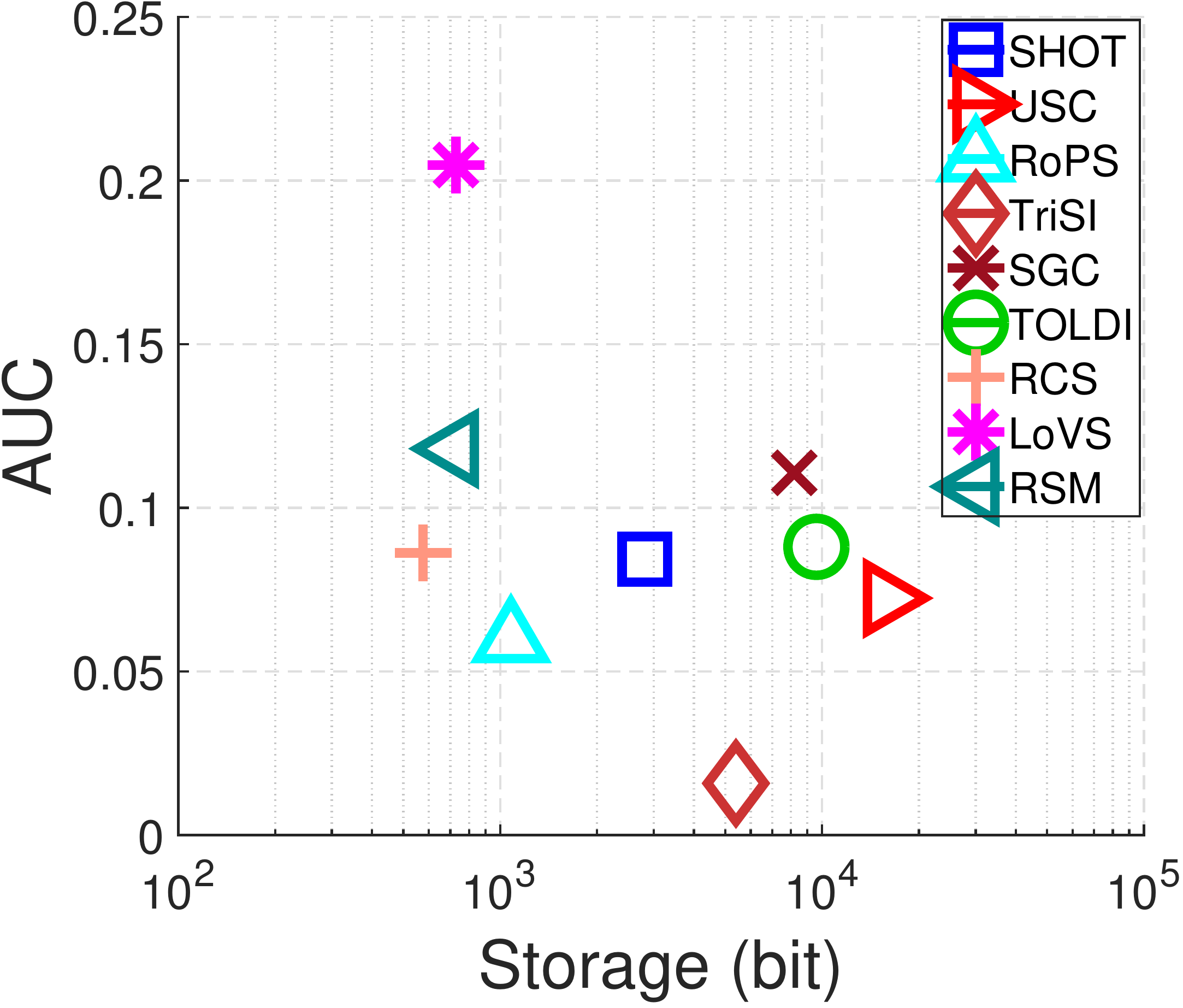}}
	\end{minipage}	
	\hfill
	\caption{Compactness performance (AUC versus storage) of evaluated feature representations on six experimental datasets.}
	\label{fig:compact}
\end{figure}
\subsection{Computational Efficiency}
\begin{figure}[t]
	\centering
	\includegraphics[width=0.85\linewidth]{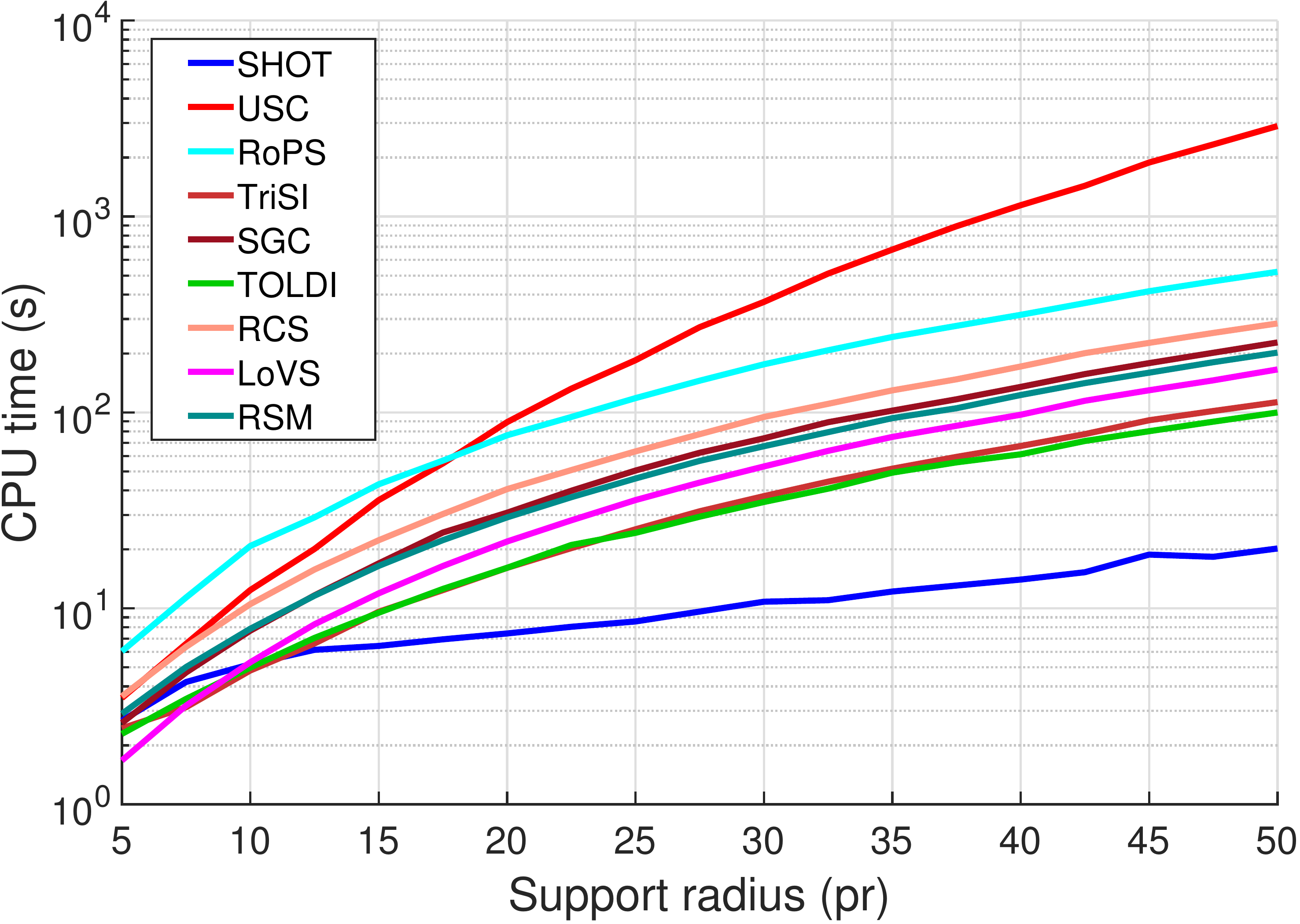}\\
	\caption{Time costs of calculating evaluated feature representations  for 1000 local surface patches with respect to different support radii.}
	\label{fig:efficiency}
\end{figure}
We assess the computational efficiency of a feature representation as follows. First, 1000 local surface patches are randomly sampled from an experimental dataset. Note that the computational efficiency of a feature representation is only related to the point count of a local surface rather than its shape geometry, so we experiment with the the \textit{Retrieval} dataset. Then, we compute the features to be tested at various scales on these sampled points. Finally, the former two steps are repeated for ten rounds and the average results are kept for evaluation. The results are shown in Fig.~\ref{fig:efficiency}.

Obviously, USC is the most time-consuming feature representation, especially for large-scale local patches. It is because USC assigns a \textit{density} value to each point in the local volume, which needs to estimate the number of points within a spherical region centered at the query point. Usually, this is done by building a $k$d-tree in the local surface to achieve fast radius neighbor search yet has a computational complexity of  $O(n{\rm log}n)$. Another observation is that features leveraging the \textit{rotation and projection} mechanism, e.g., RoPS and RCS, are  also inefficient to calculate. The reason is that rotating the local surface requires matrix multiplication, which has a non-linear computational complexity. SHOT is the most efficient feature when the support radius is greater than 12.5 \textit{pr}, followed by TOLDI and SGC. The high efficiency of SHOT is benefited from the low computational complexity of attribute extraction (i.e., the cosine of two unit vectors) and histogram binning. Since local geometric feature matching usually employs features with a scale around 15 \textit{pr}~\cite{guo2016comprehensive,yang2017eval}, SHOT, TOLDI, SGC, and LoVS are recommended choices.  
\section{Summary and Discussion}\label{sec:sum}
Based upon the evaluation results in Sect.~\ref{sec:result}, it is necessary to give a summary of the performance of evaluated features in different assessing scenarios, in order to  highlight the advantages and limitations of these feature representations to guide real-world applications and new descriptor crafting. Besides, we discuss about some new findings based on our evaluation that may indicate  future research directions in the research field of local geometric feature description.
\subsection{Performance Summary}
\begin{table}[t]\scriptsize
	\centering
	\caption{Superior and inferior feature representations summarized based on the outcomes in Sect.~\ref{sec:result}.}
	\label{tab:performance_sum}
	\scalebox{1}{
		\begin{tabular}{|c| c|c|c|}
			\hline
			\multicolumn{2}{|c|}{}& \bf Superior features & \bf Inferior features\\
			\hline
			\multirow{6}{*}{\rotatebox{90}{\bf Distinctiveness}}&\textit{Retrieval}&SGC, LoVS&SHOT\\
			\cline{2-4}
			&\textit{Laser Scanner}&LoVS, SGC&SHOT, RoPS\\
			\cline{2-4}
			&\textit{Kinect}&SGC, LoVS, SHOT&TriSI, USC, RCS\\
			\cline{2-4}
			&\textit{Space Time}&SGC, SHOT, LoVS&RCS, USC\\
			\cline{2-4}
			&\textit{LiDAR Registration}&LoVS&RoPS, TriSI\\
			\cline{2-4}
			&\textit{Kinect Registration}&LoVS&TriSI\\
			\hline
			\multirow{11}{*}{\rotatebox{90}{\bf Robustness}}&Varying support radii&LoVS, SGC, RSM& RoPS, SHOT, RCS\\
			\cline{2-4}
			&Distance to boundary&LoVS, SGC&RoPS, RCS, SHOT\\
			\cline{2-4}
			&Clutter&LoVS, SGC&RoPS, SHOT\\
			\cline{2-4}
			&Occlusion&LoVS, SGC&RoPS, SHOT\\
			\cline{2-4}
			&Partial overlap&LoVS&RoPS\\
			\cline{2-4}
			&Gaussian noise&SGC&USC, RCS\\
			\cline{2-4}
			&Shot noise&TriSI, RSM, TOLDI&USC\\
			\cline{2-4}
			&Uniform data decimation&LoVS, SGC&USC, TriSI\\
			\cline{2-4}
			&Random data decimation&LoVS, SGC&TriSI, RCS\\
			\cline{2-4}
			&Keypoint localization error&RoPS, SHOT&TriSI, USC\\
			\cline{2-4}
			&LRF error&SGC, LoVS, RoPS& SHOT, RCS, TriSI\\
			\cline{2-4}
			\hline
			\multirow{6}{*}{\rotatebox{90}{\bf Compactness}}&\textit{Retrieval}&LoVS, RSM, RoPS&USC, TOLDI\\
			\cline{2-4}
			&\textit{Laser Scanner}&LoVS&USC, TOLDI\\
			\cline{2-4}
			&\textit{Kinect}&LoVS, RSM, RoPS&USC\\
			\cline{2-4}
			&\textit{Space Time}&LoVS&USC\\
			\cline{2-4}
			&\textit{LiDAR Registration}&LoVS&USC, SGC, TOLDI\\
			\cline{2-4}
			&\textit{Kinect Registration}&LoVS&USC, SGC, TOLDI\\
			\cline{2-4}
			\hline
			\multicolumn{2}{|c|}{\bf Efficiency}&SHOT & USC, RoPS \\
			\hline
	\end{tabular}}
\end{table}
A performance summary of evaluated feature representations based on the results in Sect.~\ref{sec:result} is presented in Table~\ref{tab:performance_sum}. Two main observations can be made. (i) Most of evaluated feature representations fail to achieve a good balance among distinctiveness, robustness, compactness, and computational efficiency. However, two features, i.e., LoVS and SGC, exhibit favorable performance in most cases. In particular, LoVS achieves top-ranked performance on most datasets, e.g., \textit{Laser Scanner}, \textit{Kinect Registration}, and \textit{LiDAR Registration}, and is robust to a variety of nuisances such as clutter, occlusion, data decimation, and LRF errors. In addition, the binary format of LoVS makes it compact as well. (ii) The evaluation results find that some features designed for a particular application, e.g., RoPS and TriSI for 3D object recognition, are even inferior to many other competitors. By contrast, there exist several features that generalize well across different applications, e.g., LoVS. It is therefore necessary to rethink the rationality of some feature description approaches for a particular application. We can also get inspirations from LoVS regarding the issue of generalization ability as it remains a very challenging task in this field~\cite{guo2016comprehensive}.
\subsection{Findings}
This evaluation presents the following interesting findings that worth future investigation.

(i) \textbf{Cubic volume vs. Spheric volume}. The majority of existing local geometric features are extracted within a cubic volume, originally to promise that corresponding surface patches can be cropped identically around the keypoint regardless object poses. With the help of LRF, a cubic volume can also achieve such goal~\cite{Quan2018Local}. As revealed by the results, features computed within a cubic volume, i.e., SGC and LoVS, generally outperform those spheric volume-based features. This is potentially because more uniform partition can be performed with a cubic volume, while irregular partition may lead to subspaces with varying sizes and therefore   complicates the feature extraction for each subspace. 

(ii) \textbf{3D-to-2D projection vs. 3D preserving}. 3D-to-2D projection is widely employed in many existing feature representations, e.g., RoPS, TriSI, RCS, and TOLDI. The motivation behind is to convert spatially irregular point clouds to regular images, in order to facilitate feature description with relatively matured image description techniques. There are also 3D preserving features such as SGC, USC, and LoVS, which first split the local 3D volume into a set of subspaces and then perform feature description. The evaluation results suggest that features based on 3D-to-2D projection are generally less distinctive than 3D preserving-based feature due to the information loss after projection. Besides stronger robustness to outliers, projection-based features are also more susceptible to other nuisances such as Gaussian noise, data decimation, clutter, and occlusion.

(iii) \textbf{Space partition}. To encode spatial information, performing space partition is a simple yet very effective way~\cite{yang2017eval}. Nonetheless, determining the number of partitions needs a special care. Less partitions can hardly fully leverage the spatial information, e.g., SHOT has 32 partitions and shows limited discriminative power; more partitions will increase the sensitivity to point-level perturbations such as noise and data decimation, e.g., USC has 1980 partitions and is very sensitive to common nuisances. LoVS and SGC seem to have attained a well balance, which have 729 and 512 partitions, respectively.

(iv) \textbf{Rethinking attribute description for subspaces}. It is a common practice to perform attribute description for subspaces in both 3D-to-2D projection-based and 3D preserving-based features. For instance, RoPS and TriSI use point density feature to describe each 2D gird after projection; SHOT and SGC respectively resort to normal deviation histograms and geometric centroids for 3D subspace description. However, now we should rethink the reasonability and necessity of such operation. Because the evaluation results demonstrate that directly replacing attribute description with occupancy labels can achieve comparative or even stronger discriminative power and more robustness to common perturbations, yet being more compact and efficient. A typical example is LoVS that simply uses structural information without attribute description for subspaces, but it is the overall best feature showed by this evaluation. In addition, RSM, as a variant of RoPS without employing density features, shows stronger robustness to shot noise, clutter, and occlusion.
\section{Conclusions}\label{sec:conc}
This paper has presented a comprehensive evaluation for several state-of-the-art local geometric feature representations, presenting complementary perspectives to existing evaluations of overall feature descriptors~\cite{guo2016comprehensive} and LRFs~\cite{yang2018toward}. The experiments were conducted on datasets addressing different application scenarios with various data modalities and nuisances such as clutter, occlusion, partial overlap, Gaussian noise, shot noise, and data decimation. The tested terms for a feature representation include distinctiveness, robustness, compactness, and computational efficiency. Based upon evaluation outcomes, a summary of the overall performance and peculiarities of evaluated features has been presented, being helpful for either devising new feature descriptors or choosing a proper feature representation in a particular application. We have also discussed some interesting findings through this evaluation that may be instructive to the following researchers.

\section*{Acknowledgments}
The authors would like to thank the publishers of datasets used in our evaluation for making them publicly available. 
\bibliographystyle{IEEEtran}
\bibliography{mybibfile}
	
\end{document}